\RequirePackage{fix-cm}
\documentclass[twocolumn]{svjour3}
\smartqed{}
\usepackage{mathptmx}
\usepackage[utf8]{inputenc}
\usepackage[T1]{fontenc}
\usepackage[UKenglish]{babel}
\usepackage{csquotes}
\usepackage{amsmath}
\usepackage{amssymb}
\usepackage{mathtools}
\usepackage{array}
\usepackage{booktabs}
\usepackage{authblk}
\usepackage[acronym]{glossaries}
\usepackage[algoruled, english]{algorithm2e}
\usepackage{graphicx}
\usepackage{enumitem}
\usepackage{url}
\usepackage[sort]{cite}
\usepackage[misc]{ifsym}
\title{Optimisation of photometric stereo methods\\
by non-convex variational minimisation
}
\titlerunning{Non-convex variational photometric stereo}
\author{Georg Radow \and Laurent Hoeltgen \and Yvain Quéau \and Michael Breuß
}
\authorrunning{Radow~\etal}
\institute{G.\ Radow \and L.\ Hoeltgen \and M.\ Breuß
  \at Chair for Applied Mathematics, BTU Cottbus-Senftenberg, Cottbus, Germany.
  email:~\url{{radow,hoeltgen,breuss}@b-tu.de}
  \and
  Y. Quéau
  \at Computer Vision Group, Technical University Munich, Garching, Germany.
  email:~\url{yvain.queau@tum.de}
}
\journalname{}
\date{Received: date / Accepted: date}
\makeatletter
\providecommand*{\diff}{\@ifnextchar^{\DIfF}{\DIfF^{}}}%
\def\DIfF^#1{%
	\mathop{\mathrm{\mathstrut d}}%
	\nolimits^{#1}\gobblespace}
\def\gobblespace{%
	\futurelet\diffarg\opspace}%
\def\opspace{%
	\let\DiffSpace\!%
	\ifx\diffarg(%
	\let\DiffSpace\relax%
	\else%
	\ifx\diffarg\[%
	\let\DiffSpace\relax%
	\else%
	\ifx\diffarg\{%
	\let\DiffSpace\relax%
	\fi\fi\fi\DiffSpace}%
\makeatother
\DeclarePairedDelimiterX{\abs}[1]{\lvert}{\rvert}{%
	\ifblank{#1}{\:\cdot\:}{#1}%
}
\DeclarePairedDelimiterX{\scprod}[2]{\langle}{\rangle}{%
	\ifblank{#1}{\:\cdot\:}{#1}%
	\,,\mathopen{}%
	\ifblank{#2}{\:\cdot\:}{#2}%
}
\DeclarePairedDelimiterX{\norm}[1]{\lVert}{\rVert}{%
	\ifblank{#1}{\:\cdot\:}{#1}
}
\DeclarePairedDelimiterX{\setc}[2]{\lbrace}{\rbrace}{%
	#1\ifblank{#2}{}{\,\delimsize\vert\,\mathopen{}#2}
}

\newcommand{\SetSymbol}[1][]{%
	\nonscript\:#1\vert\allowbreak\nonscript\:\mathopen{}}
\DeclarePairedDelimiterX{\set}[1]{\lbrace}{\rbrace}{%
	
	#1
}
\DeclareMathOperator*{\argmin}{arg\,min}

\DeclareMathOperator{\prox}{prox}
\DeclareMathOperator{\vecop}{vec}

\newcommand{\Real}{\mathbb{R}}
\newcommand{\etal}{\emph{et al.}}
\newcommand{\ie}{\emph{i.e.}}
\newcommand{\cf}{\emph{c.f.}}
\newcommand{\eg}{\emph{e.g.}}

\newacronym{ML}{ML}{maximum likelihood}
\newacronym{PS}{PS}{photometric stereo}
\newacronym{SFS}{SFS}{shape from shading}
\newacronym{IIE}{IIE}{image irradiance equation}
\newacronym{FBS}{FBS}{forward-backward splitting}
\newacronym{BC}{BC}{boundary conditions}
\newacronym{BLUE}{BLUE}{best linear unbiased estimate}
\newacronym{PDE}{PDE}{partial differential equation}
\newacronym{SVD}{SVD}{singular value decomposition}
\newacronym{DCT}{DCT}{discrete cosine transform}
\newacronym{MAE}{MAE}{mean angular error}
\SetAlFnt{\small}
\glsdisablehyper
\begin{document}
%
\maketitle
\begin{abstract}
      Estimating shape and appearance of a three dimensional object from a given
      set of images is a classic research topic that is still actively pursued.
      Among the various techniques available, \glsentrylong{PS} is distinguished
      by the assumption that the underlying input images are taken from the same
      point of view but under different lighting conditions. The most common
      techniques provide the shape information in terms of surface normals. In
      this work, we instead propose to minimise a much more natural objective 
      function, namely the reprojection error in terms of depth.
      Minimising the resulting non-trivial variational model for
      \glsentrylong{PS} allows to recover the depth of the photographed
      scene directly. As a solving strategy, we follow an approach based on a
      recently published optimisation scheme for non-convex and non-smooth cost functions.\par 
      The main contributions of our paper are of theoretical nature. A
      technical novelty in our framework is the usage of matrix differential
      calculus. We supplement our approach by a detailed convergence analysis of the resulting
      optimisation algorithm and discuss possibilities to ease the computational 
      complexity. At hand of an experimental evaluation we
      discuss important properties of the method. Overall, our strategy 
      achieves more accurate results than competing
      approaches. The experiments also highlights some practical aspects of the 
      underlying optimisation algorithm 
      that may be of interest in a more general context. 
\end{abstract}
%
\section{Introduction}
\label{sec:introduction}
%
The reconstruction of three dimensional depth information given a set of two
dimensional input images is a classic problem in computer vision. The class 
of methods fulfilling this task by inferring local shape from brightness 
analysis is called photometric methods~\cite{BKP1986,Woehler2013}. They usually
 employ a static view
point and variations in illumination to obtain the 3D structure. Fundamental
photometric reconstruction processes are \gls{SFS} and \gls{PS}~\cite{BKP1986}.
\Glsentrylong{SFS} typically requires a single input image, whereas PS makes use
of several input images taken from a fixed view point under different
illumination. \Glsentrylong{PS} incorporates \gls{SFS} in the sense that
\gls{SFS} equations applied to each of the input images are integrated into a
common \gls{PS} process in order to obtain the 3D shape. This integrated model
is usually formulated as an optimisation task that best explains the input
images in terms of a pointwise estimation of shape and appearance.\par
The pioneer of the \gls{PS} method was Woodham in 1978
\cite{Woodham1978a}, see also Horn~\etal~\cite{HWS78}. The
mathematical formulation of the PS problem is based on the use of the \gls{IIE}
as in \gls{SFS} for the individual input images, respectively. The
\glsentrylong{IIE} constitutes a relation between the image intensity and the
reflectance map. The classic proceeding is thereby to consider Lambert's law
\cite{Lambert1760} for modelling the appearance of a shape given information on
its geometry and albedo as well as the lighting in a scene. It has been shown
that the orientation of a Lambertian surface can be uniquely determined from the
resulting appearance variations provided that the surface is illuminated by at
least three known, non-coplanar light sources, corresponding to at least three
input images~\cite{Woodham1980}. However, let us also mention the classic work
of Kozera~\cite{Kozera91} as well as Onn and Bruckstein~\cite{OB90} where
refined existence and uniqueness results are presented for the two-image case. 
As a beneficial aspect
beyond the possible estimation of 3D shape, PS enables to compute an albedo map
allowing to deal with non-uniform object materials or textured objects in a
photographed scene.\par
As to complete our brief review of some general aspects of PS, let us note that
it is possible to extend Woodham's classic PS model as for instance to
non-Lambertian reflectance as 
\eg~in~\cite{Bartal2017,Ikehata2014a,KhBoBr17,Mecca2016,TMDD2016}, or to take
into account several types of lighting in a scene, see
\eg~\cite{BJK07,QuDuWuCrLaDu17}. One may also consider a \gls{PS}
approach based on solving partial differential equations (PDEs) corresponding to
ratios of the underlying \gls{IIE}s, see for instance~\cite{MeRoCr15,TMDD2016}.
The latter approach makes it possible to compute the 3D shape directly whereas
in most methods following the classic PS setting a field of surface
normals is computed which needs to be integrated in another step; see
\eg~\cite{BBQBD17} for a recent discussion of integration techniques.\par
Let us turn to the formulation of the \gls{PS} approach we make use of. At this
stage we keep the presentation rather general as we elaborate on the details
that are of some importance in the context of applying our optimisation approach
in Section~\ref{section-construction}. However, in order to explain the
developments documented in this paper, it is useful to provide some formulae
here.\par
\begin{figure*}
      \centering
      \begin{tabular}{ccc}
        \includegraphics[width = 0.32\textwidth]{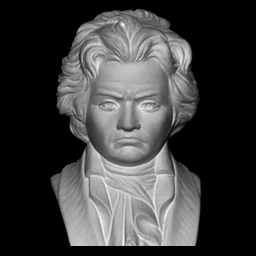} &
        \includegraphics[width = 0.32\textwidth]{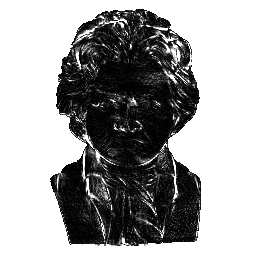} &
        \includegraphics[width = 0.32\textwidth]{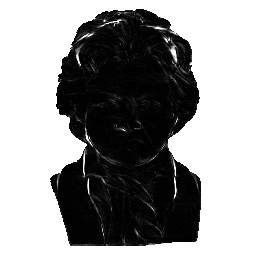} \\
        \parbox[t]{0.32\textwidth}{
        \centering{}Example input image for \gls{PS}} &
        \parbox[t]{0.32\textwidth}{
        \centering{}Classic \gls{PS} with integration} &
        \parbox[t]{0.32\textwidth}{
        \centering{}Our method}
      \end{tabular}\vspace{1.0ex}
      \caption{From a set of $m \geq 3$ images (\cf~left), classic \gls{PS}
        provides an albedo and a normal map which best explain the input images
        in the sense of a local, pointwise estimation. In a second step, the
        smooth depth map is estimated by integration. Yet, the final surface is
        not the best explanation of the images, as indicated by the reprojection
        error (\cf~energy in~\eqref{eq:reproj_error_global}) (middle). We
        display this using white for $2.5\cdot{}10^{-3}$ and black for zero.
        Instead of this local procedure, we propose to minimise the reprojection
        criterion in terms of the depth and the albedo, through global
        non-convex optimisation. Not only the images are better explained
        (right), but we also demonstrate that the 3D-reconstruction results are
        improved (\cf~Section~\ref{sec:numerical-evaluation}).}
      \label{fig:1}
\end{figure*}

Photometric 3D reconstruction is often formulated as an inverse problem: 
given an image $I$, the aim is to compute a
depth map $z$ that best explains the observed grey levels of the data. To this
end we use the \gls{IIE} $I(u,v) = \mathcal{R}(z(u,v); \vec{s},\rho(u,v))$ where
$\left(u, v\right) \in \Omega$ represent the coordinates over the reconstruction
domain $\Omega \subseteq \Real^{2}$ and where $\mathcal{R}$ denotes the
reflectance map~\cite{BKP1986}. This model describes interactions between the
surface $z$ and the lighting $\vec{s}$. The vector $\rho$ represents reflectance
parameters as \eg~the albedo, which can be either known or considered as
hidden unknown parameters. For the sake of simplicity, we will consider in this
paper only Lambertian reflectance without shadows, and we assume that the
lighting of a photographed scene is directional and known. Moreover our camera
is assumed to perform an orthographic projection. As in \gls{PS} several input
images $I^i$, $i\in\{1,\dots,m\}$ are considered under varying lighting
$\vec{s}^i$, $i\in\{1,\dots,m\}$, the \gls{PS} problem consists in finding a
depth map $z$ that best explains all \glspl{IIE} simultaneously:
\begin{equation}
\label{eq:irradiance_equation_PS}
I^{i}(u, v) = \mathcal{R}(z(u,v); \vec{s}^i,\rho),\qquad i \in \{1, \ldots, m\}
\end{equation}
\textbf{Our contribution.} Our aim is to obtain the solution $z$ of the \gls{PS}
problem, see also Figure~\ref{fig:1} for an account. We show that estimating the
optimal solution necessarily involves non-trivial optimisation methods, even
with the simplest models for the reflectance function $\mathcal{R}$ and the most
simple deviations from the model assumptions that may occur, \ie~we consider
Lambertian reflectance without shadows and additive, zero-mean Gaussian noise.\par
To achieve our goal we propose a numerical framework to approximate an optimal
solution which can be used to refine classic \gls{PS} results. Our approach
relies on matrix differential theory for analytic derivations and on recent
developments in non-convex optimisation. In that novel framework for this class
of problems we prove here the convergence of the optimisation method. The
theoretical results are supplemented by a thorough numerical investigation that
highlights some important observations on the optimisation routine.\par
The basic procedure of this work has been the subject of our conference paper
\cite{HoQuBrRa16}, the results of which are mainly contained in the second, third and
beginning of the fourth section of this article. Our current paper extends that
previous work significantly by providing the mathematical validation of
convergence and the extended analysis of the numerical optimisation algorithm.
These are also exactly the core contributions of this paper. Moreover, we give a
much more detailed description of the matrix calculus framework we employ.
%
\section{Construction of our method and more related work}
\label{section-construction}
%
As shown by Woodham~\cite{Woodham1980}, all surface normals can be estimated in
the classic \gls{PS} model without ambiguity, provided $m \geq 3$ input images
and non-coplanar calibrated lighting vectors are given. In addition, the reflectance
parameters (\eg~the albedo) can also be estimated. This is usually achieved by
minimising the difference between the given data, \ie~the input images and the
reprojection according to the estimated normal and albedo:
\begin{equation}
      \argmin_{\vec{n},\rho}\frac{1}{m} \iint_{\Omega} \sum_{i=1}^m \Phi
	\left( I^i - \mathcal{R}\left(\vec{n}; \vec{s}^i, \rho \right) \right)
        \diff{u}\diff{v}
        \label{eq:reproj_error}
\end{equation}
with a penaliser $\Phi$. As a result, one obtains an approximation of the normal
$\vec{n}(u,v)$ and the albedo $\rho(u,v)$ at each position $(u,v)$.\par
Since there is no coupling between the normals estimated in two neighboring pixels, 
those estimates are the optimal \emph{local} explanations of the image, in
the sense of the estimator $\Phi$. Yet, the estimated normal field is in general
not integrable. Thus, the depth map that can be obtained by integration is not
an optimal image explanation, but only a smooth explanation of the noisy normal
field, \cf~Figure~\ref{fig:1}.\par
Instead of this pointwise joint estimation of the normal and the albedo, it is,
as already mentioned in the introduction, possible to employ photometric ratios.
Following that procedure means to divide the $i$-th by the $j$-th \gls{IIE}
in~\eqref{eq:irradiance_equation_PS}. This way, one obtains a homogeneous linear
system in each normal vector that does not depend on the albedo,
see~\cite{MeRoCr15}. However, these ratios introduce additional difficulties in
the models. It is common to assume that image data is corrupted by additive,
zero-mean, Gaussian noise. In that case the \gls{ML} function should be chosen
to be quadratic. Unfortunately, the ratio of two Gaussian random variables
follows a Cauchy distribution~\cite{Hinkley1969}. Thus, additional care has to
be taken to find the most efficient penaliser. Another frequent assumption is
that the estimated normal fields should be integrable, yet, this is a rather
restrictive assumption. The normal field computed by many aforementioned
\gls{PS} approaches does not necessarily need to be integrable. Hence, the
integration task is usually formulated as another optimisation problem which
aims at minimising the discrepancy between the estimated normal field and that
of the recovered surface. Following that approach we now go into some more
details.\par
Assuming orthographic camera projection, the relation between the normal
$\vec{n}(u,v)$ and the depth $z(u,v)$ is given by:
\begin{equation}
      \label{eq:def_normal}
      \vec{n}(u,v) \coloneqq \frac{1}{\sqrt{\|\nabla z(u,v)\|^2 +1}}
      \left[-\nabla z(u,v),\,1 \right]^\top
\end{equation}
where $\nabla z$ is the gradient of $z$. Then, the best smooth surface
explaining the computed normals can be estimated in several ways~\cite{BBQBD17},
for instance by solving the variational problem:
\begin{equation}
      \argmin_{z} \iint_{\Omega} \Psi\left( \left\| \nabla z +
                  \begin{bmatrix}
                        \vec{n}_1/\vec{n}_3 \\
                        \vec{n}_2/\vec{n}_3
                  \end{bmatrix} \right\|^2 \right) \diff{u}\diff{v}
      \label{eq:integ_error}
\end{equation}
where $\Psi$ is again some estimator function; see~\cite{Durou2009,Harker2014}
for some discussion.\par
One may realise that, at this stage of the process chain of \gls{PS} with
integration, the images are not explicitly considered anymore. Thus, the final
surface is in general not necessarily optimal in the sense of the reprojection
criterion. Regularising the normal field before integration
\cite{Reddy2009,Zeisl2014} may also ensure integrability, but since such methods
only use the normal field, and not the images, they may be unable to assert
optimality with respect to the reprojection.\par
\emph{Global \gls{PS} approaches} solve the latter problem as they represent a
way to ensure that the recovered surface is optimal with respect to the
reprojection criterion. Moreover, it is possible to solve the
system~\eqref{eq:irradiance_equation_PS} directly in terms of the depth
\cite{Clark1992}: this ensures both that the recovered surface is regular, and
that it is optimal with respect to the reprojection criterion, calculated from
the depth map $z$ and not from a non-integrable estimate of its gradient. Some
PDE-based PS approaches have been recently proposed, and were shown to ease the
resolution in particularly difficult situations such as pointwise lighting~\cite{QuDuWuCrLaDu17} 
and specular reflectance~\cite{TMDD2016}. To
ensure robustness, such methods can be coupled with variational 
methods. In other words, the criterion which should be
considered for ensuring optimality of a surface reconstruction by \gls{PS} is
not the local criterion~\eqref{eq:reproj_error}, but rather:
\begin{equation}
      \argmin_{z,\rho}
	\frac{1}{m} \iint_{\Omega}
	\sum_{i=1}^m \Phi\left( I^i - \mathcal{R}\left(z; \vec{s}^i,\rho\right) \right)
        \diff{u}\diff{v}
        \label{eq:reproj_error_global}
\end{equation}
A theoretical analysis of the choice $\Phi(x) = |x|$, can be found
in~\cite{Chabrowski1993}. Numerical resolution methods based on proximal
splittings were more recently introduced in~\cite{SSVM2015a}. Yet, this last
work relies on an ``optimise then discretise'' approach which would involve
non-trivial oblique \gls{BC}, replaced there for simplicity reasons by Dirichlet
\gls{BC}. Obviously, this represents a strong limitation which prevents working
with many real-world data where this oblique \gls{BC} is rarely available.\par
The optimisation problem~\eqref{eq:reproj_error_global} is usually non-linear
and non-convex. The ratio procedure described earlier can be used: it
simultaneously eliminates the albedo and the non-linear terms,
\cf~\cite{Smith2016,Mecca2016,TMDD2016,GSSM2015} and obviously removes the bias
due to non-integrability. But let us recall that it is only the best linear unbiased estimate, and
also not the optimal one. To guarantee optimality, it is necessary to minimise
the nonlinear, non-convex energy, \ie~without employing ratios. Other 
methods~\cite{QuDuWuCrLaDu17,Queau2017} overcome the nonlinearity by absorbing 
it in the auxiliary albedo variable. Again, the solution is not that of the original 
problem~\eqref{eq:reproj_error_global} which remains, to the best of our knowledge, 
unsolved.\par
Solving~\eqref{eq:reproj_error_global} is a challenging problem. Efficient
strategies to find the sought minimum are scarce. Recently Ochs~\etal~\cite{OCBP2013} 
proposed a novel method to handle such non-convex
optimisation problems, called \emph{iPiano}. A major asset of the approach is
the extensive convergence theory provided in~\cite{OCBP2013,P2016}. Because of
this solid mathematical foundation we explore the iPiano approach in this work.
The scheme makes explicit use of the derivative of the cost function, which in
our case involves derivatives of matrix-valued functions, and we will employ as
a technical novelty, matrix differential theory~\cite{MN1985,MN2007} to derive
the resulting scheme.
%
\section{Non-convex discrete variational model for PS}
\label{sec:non-conv-vari}
%
In this section we describe the details of our framework for estimating both the
depth and the (Lambertian) reflectance parameters over the domain $\Omega$.
%
\subsection{Assumptions on the PS model}
\label{sec:assumptions}
%
We assume $m \geq 3$ grey level images $I^i$, $i \in \{1, \dots, m\}$, are
available, along with the $m$ lighting vectors $\vec{s}^{i} \in \Real^3$,
assumed to be known and non-coplanar. We also assume Lambertian reflectance and
neglect shadows, which leads to the following well-known model:
\begin{equation}
      \mathcal{R}\left(\vec{n}(u,v); \vec{s}^i,\rho\right)
      \coloneqq
      \rho\left( u, v\right) \scprod*{\vec{s}^i}{\vec{n}\left( u, v\right)}
      \label{eq:lambertian-model-ps}
\end{equation}
where $(u,v) \in \Omega$, $i = 1, \ldots, m$ and $\rho(u,v)$ is the albedo at
the surface point conjugated to position $(u,v)$, considered as a hidden unknown
parameter. Let us note that real-world \gls{PS} images can be processed by
low-rank factorisation techniques in order to match the linear reflectance model
\eqref{eq:lambertian-model-ps}, \cf~\cite{Wu2010}.\par
We further assume orthographic projection, hence the normal $\vec{n}(u,v)$ is
given by~\eqref{eq:def_normal}. Then the reflectance model becomes a function of
the depth map $z$:
\begin{multline}
      \mathcal{R}\left(z; \vec{s}^i,\rho\right) \coloneqq
      \frac{\rho\left( u, v\right)}{\sqrt{\|\nabla z(u,v)\|^2+1}}
      \scprod*{\vec{s}^{i}}{\begin{bmatrix}-\nabla z(u, v) \\1\end{bmatrix}}
\end{multline}
with $(u,v) \in \Omega$, for all $i$. Eventually, we assume that the images $I^i$
differ from this reflectance model only up to additive, zero-mean, Gaussian
noise. The \gls{ML} estimator is thus the least-squares estimator
$\Phi(x)=\frac{1}{2} x^2$, and the cost function in the reprojection
criterion~\eqref{eq:reproj_error_global} becomes:
\begin{multline}
      \mathcal{E}_{\mathcal{R}} \left(z,\rho ; I\right) \coloneqq
      \frac{1}{2m} \iint_{\Omega} \sum_{i=1}^m
      \biggl( I^i - \mathcal{R}\left(z; \vec{s}^i,\rho\right)
      \biggr)^2 \diff{u}\diff{v}
      \label{eq:PS_var}
\end{multline}
%
\subsection{Tikhonov regularisation of the model}
\label{sec:tikh-regul-model}
%
Our energy in~\eqref{eq:PS_var} only depends on the gradient $\nabla z$ and not
on the depth $z(u,v)$ itself. As a consequence, solutions can only be determined
up to an arbitrary constant. As a remedy we follow~\cite{Mecca2016} and introduce
a reference depth $z_{0}(u,v)$, thus regularising our initial model with a
zero-th order Tikhonov regulariser controlled by a parameter $\lambda > 0$:
\begin{equation}
      \label{eq:PS_var_reg}
      \argmin_{z,\rho}
      \mathcal{E}_{\mathcal{R}}\left( z,\rho ; I\right)  +
      \frac{\lambda}{2}
      \iint_{\Omega} \left( z - z_0 \right)^2
      \diff{u}\diff{v}
\end{equation}
In practice, $\lambda$ can be set to any small value, so that a solution
of~\eqref{eq:PS_var_reg} lies as close as possible to a minimiser
of~\eqref{eq:PS_var}. In all our experiments we set $\lambda := 10^{-6}$ and
$z_0$ as the classic \gls{PS} solution followed by least-squares
integration~\cite{BBQBD17}.\par
%
\subsection{Discretisation}
\label{sec:discretisation}
%
As already mentioned, ``optimise then discretise'' approaches for
solving~\eqref{eq:PS_var_reg}, such as~\cite{SSVM2015a}, involve non-trivial
\gls{BC}. Hence, we prefer a ``discretise then optimise'', finite dimensional
formulation of the variational \gls{PS} problem~\eqref{eq:PS_var_reg}.\par
In our discrete setting we are given $m$ images $\vec{I}^i$, $i \in
\{1,\dots,m\}$, with $n$ pixels labelled with a single index $j$ running from 1
to $n$. We discretise~\eqref{eq:PS_var_reg} in the following way:
\begin{multline}
      \argmin_{\vec{z}, \rho \in\Real^{n}}\biggl\{\frac{1}{2m}
      \sum_{j}\biggl\|
      \vec{I}_{j} - \frac{\rho_{j}}{\sqrt{\norm*{\nabla \vec{z}_{j}}^{2} + 1}}
      \vec{S}
      \begin{bmatrix}
            -\nabla \vec{z}_{j} \\
            1
      \end{bmatrix}
      \biggr\|^{2} \\
      + \frac{ \lambda }{2} \norm*{\vec{z}_{j} - {\vec{z}_{0}}_{j}}^{2}
      \biggr\}
      \label{eq:16}
\end{multline}
where $\vec{I}_j \coloneqq [\vec{I}^1_j,\dots,\vec{I}^m_j]^\top \in
\mathbb{R}^m$ is the vector of intensities at pixel $j$, $\nabla \vec{z}_j$
represents now a finite difference approximation of the gradient of $\vec{z}$ at
pixel $j$, and $\vec{S} = [\vec{s}^1,\ldots,\vec{s}^m]^\top \in \Real^{m, 3}$ is
a matrix containing the stacked $m$ lighting vectors $\vec{s}^i$.\par
We remark that the matrix $\vec{S}$ can be decomposed into two sub-matrices
$\vec{S}_{\ell}$ and $\vec{S}_{r}$ of dimensions $m\times 2$ and $m\times 1$
such that $\vec{S}\coloneqq\begin{bmatrix}\vec{S}_{\ell} &
\vec{S}_{r}\end{bmatrix}$, and so that
\begin{equation}
      \label{eq:17}
      \vec{S}
      \begin{bmatrix}
            -\nabla \vec{z}_{j} \\
            1
      \end{bmatrix}
      = -\vec{S}_{\ell}\nabla \vec{z}_{j} + \vec{S}_{r}
\end{equation}
Let us also introduce a $2n \times n$ block matrix $\vec{M}$, such that each
block $\vec{M}_j$ is a $2 \times n$ matrix containing the finite difference
coefficients used for approximating the gradient:
\begin{equation}
      \vec{M} \coloneqq
      \begin{bmatrix}
            \vec{M}_1 \; \ldots \; \vec{M}_n
      \end{bmatrix}^\top
      \in \Real^{2n,n},
      \qquad
      \vec{M}_j \vec{z} = \nabla \vec{z}_j \in \Real^2
\end{equation}
We further introduce the aliases
\begin{equation}
      \label{eq:18}
      \vec{A}_{j} (\vec{z},\rho) \coloneqq
      -\frac{\rho_{j}}{\sqrt{1+\norm*{ \vec{M}_j \vec{z}}^{2}}} \vec{S}_{\ell} \in \Real^{m,2}
\end{equation}
and
\begin{equation}
\label{eq:18.5}
	\vec{b}_{j} (\vec{z},\rho) \coloneqq \vec{I}_{j} -
	\frac{\rho_{j}}{\sqrt{1+\norm*{ \vec{M}_j \vec{z}}^{2}}} \vec{S}_{r} \in \Real^{m}
\end{equation}
and stack them, respectively, in a block-diagonal matrix
\begin{equation}
      \vec{A}(\vec{z},\rho)
      \coloneqq
      \begin{bmatrix}
            \vec{A}_1(\vec{z},\rho) & ~      & ~ \\
            ~                       & \ddots & ~ \\
            ~                       & ~      & \vec{A}_n(\vec{z},\rho)
      \end{bmatrix}
      \in \Real^{mn,2n}
      \label{eq:definitionA}
\end{equation}
and a vector
\begin{equation}
      \vec{b}(\vec{z},\rho) \coloneqq
      \begin{bmatrix}
            \vec{b}_1(\vec{z},\rho) \\
            \vdots \\
            \vec{b}_n(\vec{z},\rho)
      \end{bmatrix}
      \in \Real^{mn}\label{eq:definitionb}
\end{equation}
Using these notational conventions as well as
\begin{equation}
      \label{eq:function_f}
      f(\vec{z},\rho) \coloneqq
      \frac{1}{2m}\norm*{
        \vec{A}(\vec{z},\rho) \vec{M}\vec{z}-\vec{b}(\vec{z},\rho)}_2^2
\end{equation}
and
\begin{equation}
      g(\vec{z}) \coloneqq
      \frac{\lambda}{2}\norm*{\vec{z}-\vec{z}_0}_{2}^{2}
\end{equation}
the task in~\eqref{eq:16} can be rewritten compactly as
\begin{equation}
      \label{eq:19}
      \argmin_{\vec{z}, \rho \in \Real^{n}}
      \set*{
        f(\vec{z},\rho) + g(\vec{z})
      }
\end{equation}
which is the discrete \gls{PS} model we propose to tackle in this paper. Observe
that, if $\vec{A}(\vec{z},\rho)$ and $\vec{b}(\vec{z},\rho)$ were constant,
problem~\eqref{eq:19} would be a linear least squares problem with respect 
to $\vec{z}$.\par
Let us remark that~\eqref{eq:19} can be easily extended to include more
realistic reflectance~\cite{Ju2013,KhBoBr17} and lighting~\cite{BMVC.28.128,QuDuWuCrLaDu17} 
models, as
well as more robust estimators~\cite{Ikehata2014a,Queau2017}: this only requires to change
the definition of $f$, which stands for the global reprojection error
$\mathcal{E}_{\mathcal{R}}$.
%
\subsection{Alternating optimisation strategy}
\label{sec:altern-optim-strat}
%
In order to ensure applicability of our method to real-world data, the albedo
$\rho$ cannot be assumed to be known. Inspired by the well-known
Expectation-Maximisation algorithm, we treat $\rho$ as a hidden parameter, and
opt for an alternating strategy which iteratively refines the depth with fixed
albedo, and the hidden parameter with fixed depth:
\begin{align}
  \label{eq:alternating}
  \vec{z}^{(k+1)} &= \argmin_{\vec{z}}
                    \set*{f\left( \vec{z},\rho^{(k)} \right) + g(\vec{z})} \\
  \rho^{(k+1)} &= \argmin_{\rho}
                 \set*{f\left( \vec{z}^{(k+1)},\rho \right) + g\left( \vec{z}^{(k+1)} \right)}
\end{align}
starting from $\vec{z}^{(0)} = \vec{z}_0$ and taking as $\rho^{(0)}$ the albedo
obtained by the classic \gls{PS} approach~\cite{Woodham1980}. Of course, the
choice of a particular prior $\vec{z}_0$ has a direct influence on the
convergence of the algorithm. The proposed scheme globally converges
towards the solution, even with a trivial prior $\vec{z}_0 \equiv
\text{constant}$, but convergence is very slow in this case. Thus the proposed
method should be considered as a post-processing technique to refine classic
\gls{PS} approaches, rather than as a standalone \gls{PS} method.\par
Now, let us comment on the two optimisation problems in~\eqref{eq:alternating}.
Updating $\rho$ amounts pointwise to a linear least-squares problem, which admits
the following closed-form solution at each pixel:
\begin{equation}
      \label{eq:rho_update}
      \rho^{(k+1)}_j =\frac{
        \sqrt{1 + \|\vec{M}_j \vec{z}^{(k+1)}\|^2 }
        \sum\limits_{i=1}^m \vec{I}^i_j {\vec{s}^i}^\top
        \begin{bmatrix}
              -\vec{M}_j \vec{z}^{(k+1)} \\
              1
        \end{bmatrix}
      }{
        \sum\limits_{i=1}^m \left( {\vec{s}^i}^\top
              \begin{bmatrix}
                    -\vec{M}_j \vec{z}^{(k+1)} \\
                    1
              \end{bmatrix}\right)^2
      }
\end{equation}
The computation of $\vec{z}^{(k+1)}$ is considerably harder, and it is dealt
with in the following paragraphs.
%
\section{An inertial proximal point algorithm for PS}
\label{sec:iPianoForPS}
%
In this section we discuss the numerical solution strategy of our problem~\eqref{eq:alternating}. 
We especially discuss the main difficulty within this strategy, that is to compute the 
gradient with respect to $\vec{z}$ for the function $f$ in~\eqref{eq:function_f}. 
Apart from the explicit formula for this gradient we also investigate an 
approximation leading to efficient computations on a desktop computer with an intel CPU.
\subsection{The iPiano algorithm}
%
We will now make precise the iPiano algorithm~\cite{OCBP2013} for our 
problem~\eqref{eq:alternating}.
Since the albedo is fixed for the purpose of the corresponding optimisation
stage, we denote $f(\vec{z}) = f( \vec{z},\rho^{(k)} )$. The iPiano algorithm
seeks a minimiser of
\begin{equation}
\label{eq:iPianoCost}
\min_{\vec{x}\in\Real^{n}}\set*{f(\vec{x}) + g(\vec{x})}
\end{equation}
where $g\colon\Real^{n}\to\Real$ is convex and $f\colon\Real^{n}\to\Real$ is
smooth. What makes iPiano appealing is the fact that $g$ must not necessarily be
smooth and $f$ is not required to be convex. This allows manifold designs of
novel fixed-point schemes. In its general form it evaluates
\begin{equation}
      \label{eq:prox-new-1}
      \prox_{\alpha g}
      \left(
            \vec{z}^{(k)} - \alpha \nabla f(\vec{z}^{(k)} ) +
            \beta (\vec{z}^{(k)} - \vec{z}^{(k-1)})
      \right)
\end{equation}
where the proximal operator is given by
\begin{equation}
      \label{eq:24}
      \prox_{\alpha g}\left( \vec{z} \right) \coloneqq
      \argmin_{\vec{x}}\set*{
        \frac{1}{2}\norm*{\vec{x}-\vec{z}}^{2} + \alpha g\left( \vec{x} \right)}
\end{equation}
and which goes back to Moreau~\cite{M1965}. Before we can define the final
algorithm we also need to determine the gradient of $f$.
%
\subsection{Matrix Calculus}
\label{sec:matrixCalculus}
%
We will first recall some general rules to derive the Jacobian of a matrix
valued function, before we apply these rules to our setting in the next
section.\par
In our setting the main difficulty is that the matrix $\vec{A}$ depends on our
sought unknown $\vec{z}$. In order to state a useful representation of arising
differential expressions we have to resort to matrix differential calculus. We
refer to~\cite{MN1985,P1985,MN2007,PP2008} for a more in-depth representation. A
key notion is the definition of the Jacobian of a matrix, which can be obtained
in several ways. In this paper we follow the one given in~\cite{MN1985}.
\begin{definition}[Jacobian of a Matrix Valued Function]
      \label{thm:MatrixJacobian}
      Let $\vec{A}$ be a differentiable $m\times p$ real matrix function of an
      $n\times q$ matrix $\vec{X}$ of real variables,
      \ie~$\vec{A}=\vec{A}(\vec{X})$. The \emph{Jacobian matrix} of $\vec{A}$
      at $\vec{X}$ is the $mp\times nq$ matrix
      \begin{equation}
            \label{eq:MatrixJacobian}
            D\left[ \vec{A} \right]\left( \vec{X} \right) \coloneqq
            \frac{\diff\vecop{\left( \vec{A}\left( \vec{X} \right) \right)}}
            {\diff(\vecop{\vec{X}})^{\top}}
      \end{equation}
      where $\vecop{(\cdot)}$ corresponds to the vectorisation operator
      described in~\cite{HJ1994} (Definition 4.29). This operator stacks
      column-wise all the entries from its matrix argument to form a large
      vector.
\end{definition}
Here, differentiability of a matrix valued function means that the corresponding
vectorised function is differentiable in the usual sense. By this definition the
computation of a matrix Jacobian can be reduced to computing a Jacobian for a
vector valued function.
\begin{example}
      \label{example:1}
      Let $\vec{A}(\vec{x})\in\Real^{m,m}$ be a differentiable diagonal matrix
      \begin{equation}
            \vec{A}(\vec{x})=
            \begin{bmatrix}
                  a_1(\vec{x}) & & \\
                  & \ddots & \\
                  & & a_m(\vec{x})
            \end{bmatrix}
            \qquad\text{for all }\vec{x}\in\Real^n
      \end{equation}
      then the Jacobian matrix of $\vec{A}$ at $\vec{x}$ has the form
      \begin{equation}
            D\left[\vec{A}(\vec{x})\right](\vec{x})=\frac{\diff}{\diff \vec{x}^\top}
            \begin{bmatrix}
                  a_1(\vec{x})\\
                  \vec{0}_{m,1}\\
                  a_2(\vec{x})\\
                  \vec{0}_{m,1}\\
                  \vdots\\
                  a_m(\vec{x})
            \end{bmatrix}
            =
            \begin{bmatrix}
                  \frac{\partial a_1(\vec{x})}{\partial \vec{x}_1} & \cdots & \frac{\partial a_1(\vec{x})}{\partial \vec{x}_n}\\
                  & \vec{0}_{m,n} & \\
                  \frac{\partial a_2(\vec{x})}{\partial \vec{x}_1} & \cdots & \frac{\partial a_2(\vec{x})}{\partial \vec{x}_n}\\
                  & \vec{0}_{m,n} & \\
                  & \vdots & \\
                  \frac{\partial a_m(\vec{x})}{\partial \vec{x}_1} & \cdots & \frac{\partial a_m(\vec{x})}{\partial \vec{x}_n}
            \end{bmatrix}
      \end{equation}
where $\vec{0}_{p,q}$ denotes a $p\times q$ block of zeros.
\end{example}\par
The following two lemmas state extensions of the product and chain-rule to
matrix valued settings. They provide us closed form representations that will be
useful for the forthcoming findings. These results have been extracted from
\cite{MN1985} (Theorem~7 and~9 respectively). Since these lemmas have been
copied verbatim, we refer to their source for the detailed proofs.
\begin{lemma}[Chain Rule]
\label{thm:a_ChainRule}
Let $S$ be a subset of $\Real^{n,q}$ and assume that $\vec{F}\colon S\to
\Real^{m,p}$ is differentiable at an interior point $\vec{C}$ of $S$. Let $T$ be
a subset of $\Real^{m,p}$ such that $\vec{F}(\vec{X})\in T$ for all $\vec{X}\in
S$, and assume that $\vec{G}\colon T\to \Real^{r,s}$ is differentiable at an
interior point $\vec{B}=\vec{F}(\vec{C})$ of $T$. Then the composite function
$\vec{H}\colon S\to \Real^{r,s}$ defined by
$\vec{H}(\vec{X})=\vec{G}(\vec{F}(\vec{X}))$ is differentiable at $\vec{C}$ and
\begin{equation}
      \label{eq:a_MatrixChainRule}
      D[\vec{H}](\vec{C}) = D[\vec{G}](\vec{B}) D[\vec{F}](\vec{C})
\end{equation}
\end{lemma}
\begin{definition}[Kronecker Product]
      \label{thm:KroneckerProduct}
      Let $\vec{A}=\left(a_{i,j}\right)$ be a $m\times n$ matrix and $\vec{B}$
      be a $p\times q$ matrix then the Kronecker Product $\vec{A}\otimes\vec{B}$
      is defined as
      \begin{equation}
            \label{eq:KroneckerProduct}
            \vec{A}\otimes\vec{B}\coloneqq
            \begin{bmatrix}
                  a_{1,1}\vec{B}& \cdots & a_{1,n}\vec{B} \\
                  \vdots & & \vdots \\
                  a_{m,1}\vec{B}& \cdots & a_{m,n}\vec{B}
            \end{bmatrix}
      \end{equation}
\end{definition}
\begin{example}
      For a row vector $\vec{B}=\left[b_1,\dots,b_n\right]\in\Real^{1,n}$ and
      the identity matrix $\vec{1}_3\in\Real^{3,3}$ we have
      \begin{equation}
            \vec{B}\otimes\vec{1}_3=
            \begin{bmatrix}
                  b_1 &     &     & b_2 &     &     &               & b_n &     & \\
                      & b_1 &     &     & b_2 &     & \cdots  \quad &     & b_n & \\
                      &     & b_1 &     &     & b_2 &               &     &     & b_n
            \end{bmatrix}
      \end{equation}
\end{example}
\begin{lemma}[Product Rule]
      \label{thm:a_MatrixProductRule}
      Let $\vec{U}\colon S\to \Real^{m,r}$ and $\vec{V}\colon S \to \Real^{r,p}$
      be two matrix functions defined and differentiable on an open set
      $S\subseteq\Real^{n,q}$. Then the matrix product $\vec{U}\vec{V}\colon S\to
      \Real^{m,p}$ is differentiable on $S$ and the Jacobian matrix
      $D[\vec{U}\vec{V}](\vec{X})\in \Real^{mp,nq}$ is given by
      \begin{equation}
            \label{eq:a_MatrixProductRule}
            D\left[ \vec{U} \vec{V} \right]\left( \vec{X} \right) = ( \vec{V}^{\top} \otimes
            \vec{1}_{m} ) D[\vec{U}](\vec{X}) + (\vec{1}_{p}\otimes \vec{U})
            D[\vec{V}](\vec{X})
      \end{equation}
      Here, $\vec{1}_{k}$ represents the identity matrix in $\Real^{k,k}$.
\end{lemma}
\begin{example}
      Let $\vec{A}$ be a differentiable $m\times m$-matrix and $\vec{M}$ be a
      $m\times n$-matrix, then by Lemma~\ref{thm:a_MatrixProductRule} we have
      \begin{equation}
            \begin{split}
                  &D\left[\vec{A}(\vec{x})\vec{M}\vec{x}\right](\vec{x})\\
                  &=\left(\left(\vec{M}\vec{x}\right)^\top\otimes\vec{1}_m\right)D\left[\vec{A}(\vec{x})\right](\vec{x})\\
                  &\quad+\left(\vec{1}_1\otimes\vec{A}(\vec{x})\right)D\left[\vec{M}\vec{x}\right](\vec{x})\\
                  &=\left(\left(\vec{M}\vec{x}\right)^\top\otimes\vec{1}_m\right)D\left[\vec{A}(\vec{x})\right](\vec{x})
                  +\vec{A}(\vec{x})\vec{M}
            \end{split}
      \end{equation}
\end{example}
\subsection{Gradient computation}
\label{sec:gradient-f}
%
The following two corollaries are a direct consequence from the foregoing
statements. It suffices to plug in the corresponding quantities. We also remind,
that our choice of the matrix derivative allows us to interpret vectors as
matrices having a single column only.
\begin{corollary}
      \label{thm:a_3}
      Let $\vec{A}(\vec{z})$ be a $n\times q$ matrix depending on
      $\vec{z}\in\Real^{m}$ and $\vec{M}\in\Real^{q,m}$ a matrix which does not
      depend on $\vec{z}$, then the Jacobian of the matrix-vector product
      $\vec{A}(\vec{z})\vec{M}\vec{z}$ is given by
      \begin{equation}
            \label{eq:a_25}
            D[\vec{A}(\vec{z})\vec{M}\vec{z}](\vec{z}) =
            \left( (\vec{M}\vec{z})^{\top} \otimes \vec{1}_{n} \right)
            D[\vec{A}](\vec{z}) + \vec{A}(\vec{z})\vec{M}
      \end{equation}
      \begin{proof}
            We apply the product rule on the product between $\vec{A}(\vec{z})$
            and $\vec{M}\vec{z}$ and subsequently on the product
            $\vec{M}\vec{z}$. In a first step this yields
            \begin{equation}
                  \label{eq:a_26}
                  D[\vec{A}(\vec{z})\vec{M}\vec{z}] =
                  \left( (\vec{M}\vec{z})^{\top}
                        \otimes \vec{1}_{n} \right) D[\vec{A}](\vec{z})
                  + \vec{A}(\vec{z}) D[\vec{M}\vec{z}](\vec{z})
            \end{equation}
            Since $D[\vec{M}\vec{z}](\vec{z})=\vec{M}$ the result follows
            immediately. \qed
      \end{proof}
\end{corollary}
Corollary~\ref{thm:a_4} and Theorem~\ref{thm:a_Gradf} yield our desired compact
representations that we use for the algorithmic presentation of our iterative
schemes.
\begin{corollary}
      \label{thm:a_4}
      Using the same assumptions as in Corollary~\ref{thm:a_3}, we deduce from
      the chain rule given in Lemma~\ref{thm:a_ChainRule} the following
      relationship
      \begin{multline}
            \label{eq:a_27}
            \nabla \norm*{\vec{A}(\vec{z}) \vec{M} \vec{z}}_{2}^{2}\\
            = 2 \bigl(
            \underbrace{
              D[\vec{A}](\vec{z})^{\top}
              \left( \left( \vec{M}\vec{z} \right) \otimes \vec{1}_{n} \right) +
              \vec{M}^{\top} \vec{A}(\vec{z})^{\top}}_{=D[\vec{A}(\vec{z})\vec{M}\vec{z}](\vec{z})^{\top}} \bigr) \vec{A}(\vec{z}) \vec{M} \vec{z}
      \end{multline}
      where $\nabla$ denotes the gradient with respect to $\mathbf{z}$.
      \begin{proof}
            Since $D\left[\norm*{\vec{x}}_{2}^{2}\right](\vec{x})$ is given by
            $2\vec{x}^{\top}$ we conclude from the chain- and product-rule that
            \begin{equation}
                  \label{eq:a_1}
                  \begin{split}
                        &D[\norm*{\vec{A}(\vec{z}) \vec{M} \vec{z}}_{2}^{2}](\vec{z})\\
                        &= 2 \bigl( \vec{A}(\vec{z}) \vec{M} \vec{z} \bigr)^{\top} D [\vec{A}(\vec{z}) \vec{M} \vec{z}] (\vec{z}) \\
                        &= 2 \bigl( \vec{A}(\vec{z}) \vec{M} \vec{z} \bigr)^{\top}\\
                        &\quad \bigl(
                        \bigl( (\vec{M}\vec{z})^{\top} \otimes \vec{1}_{n} \bigr) D[\vec{A}](\vec{z})
                        + \vec{A}(\vec{z}) D[\vec{M}\vec{z}](\vec{z}) \bigr)
                  \end{split}
            \end{equation}
            Since the gradient is simply the transposed version of the Jacobian,
            we obtain
            \begin{multline}
                  \label{eq:a_2}
                  \nabla \norm*{\vec{A}(\vec{z}) \vec{M} \vec{z}}_{2}^{2}
                  = 2 \Bigl( \left( (\vec{M}\vec{z})^{\top} \otimes \vec{1}_{n} \right) D[\vec{A}](\vec{z})\\
                  + \vec{A}(\vec{z}) D[\vec{M}\vec{z}](\vec{z}) \Bigr)^{\top}
                  \vec{A}(\vec{z}) \vec{M} \vec{z}
            \end{multline}
            from which the statement follows immediately.
            \qed
      \end{proof}
\end{corollary}
Let us now come to our main result.
\begin{theorem}
      \label{thm:a_Gradf}
      Let $f(\vec{z}) = \norm*{\vec{A}(\vec{z})\vec{M}\vec{z}
        -\vec{b}(\vec{z})}_{2}^{2}$ be given with sufficiently smooth data
      $\vec{A}(\vec{z})$ and $\vec{b}(\vec{z})$. Then we have for the gradient of $f$
      the following closed form expression:
      \begin{multline}
            \nabla f(\vec{z})  = \, 2 \left(\vec{A}(\vec{z})\vec{M} +
                  \left( (\vec{M}\vec{z})^{\top} \otimes \vec{1}_{n} \right)  D[\vec{A}](\vec{z}) -
                  D[\vec{b}](\vec{z}) \right)^{\top} \\
            \, (\vec{A}(\vec{z})\vec{M}\vec{z} - \vec{b}(\vec{z}))
            \label{eq:a_grad_f}
      \end{multline}
      \begin{proof}
            From the relationship between the canonical scalar product in
            $\mathbb{R}^{n}$ and the Euclidean norm we deduce that
            \begin{multline}
                  \label{eq:a_29}
                  \norm*{\vec{A}(\vec{z}) \vec{M} \vec{z} - \vec{b}(\vec{z})}_{2}^{2}\\
                  =\norm*{\vec{A}(\vec{z})\vec{M}\vec{z}}_{2}^{2} + \norm*{\vec{b}(\vec{z})}_{2}^{2} -
                  2\scprod*{\vec{A}(\vec{z})\vec{M}\vec{z}}{\vec{b}(\vec{z})}
            \end{multline}
            Applying the gradient at each term separately and using the results
            from the previous corollaries, we obtain
            \begin{multline}
                  \label{eq:a_30}
                  \nabla \norm*{\vec{A}(\vec{z})\vec{M}\vec{z}-\vec{b}(\vec{z})}_{2}^{2} \\
                  = 2\bigl( \underbrace{ D[\vec{A}](\vec{z})^{\top}
                    \left( \left( \vec{M}\vec{z} \right) \otimes \vec{1}_{n} \right) +
                    \vec{M}^{\top} \vec{A}(\vec{z})^{\top} }_{=D[\vec{A}(\vec{z})\vec{M}\vec{z}](\vec{z})^{\top}} \bigr)
                  \vec{A}(\vec{z}) \vec{M} \vec{z}\\
                  \quad + 2 D[\vec{b}](\vec{z})^{\top} \vec{b}(\vec{z})\\
                  \quad- 2 \bigl( \underbrace{ D[\vec{A}](\vec{z})^{\top}
                    \left( \left( \vec{M}\vec{z} \right) \otimes \vec{1}_{n} \right) +
                    \vec{M}^{\top} \vec{A}(\vec{z})^{\top} }_{=D[\vec{A}(\vec{z})\vec{M}\vec{z}](\vec{z})^{\top}} \bigr) \vec{b}(\vec{z})\\
                  - 2 D[\vec{b}](\vec{z})^{\top} \vec{A}(\vec{z})\vec{M}\vec{z}
            \end{multline}
            which can be simplified to
            \begin{multline}
                  \label{eq:a_31}
                  \nabla \norm*{\vec{A}(\vec{z})\vec{M}\vec{z}-\vec{b}(\vec{z})}_{2}^{2}\\
                  = 2 (D[\vec{A}(\vec{z})\vec{M}\vec{z}](\vec{z})-D[\vec{b}](\vec{z}))^{\top}
                  (\vec{A}(\vec{z})\vec{M}\vec{z}-\vec{b}(\vec{z}))
            \end{multline}
            The result follows now from the linearity of the Jacobian.
            \qed
      \end{proof}
\end{theorem}
Now, we obtain for the gradient of the function $f$ from~\eqref{eq:function_f},
resp.~\eqref{eq:a_grad_f}:
\begin{multline}
      \nabla f(\vec{z}) = \frac{1}{m}
      \Bigl(\vec{A}(\vec{z})\vec{M} + \left( (\vec{M}\vec{z})^{\top} \otimes \vec{1}_{nm} \right)
      D[\vec{A}](\vec{z})\\
      - D[\vec{b}](\vec{z}) \Bigr)^{\top}
      (\vec{A}(\vec{z})\vec{M}\vec{z} - \vec{b}(\vec{z}))
      \label{eq:grad_f}
\end{multline}
The addition of $\left( (\vec{M}\vec{z})^{\top} \otimes \vec{1}_{nm} \right)
D[\vec{A}](\vec{z})- D[\vec{b}](\vec{z})$ stems from the inner derivative, since
$\vec{A}$ and $\vec{b}$ are not constant.
%
\subsection{Approximation of the gradient of $f$}
\label{sec:gradientApprox}
%
Our numerical scheme depends on a gradient descent step of $f$
from~\eqref{eq:function_f} (resp.~\eqref{eq:a_grad_f}) with respect to
$\vec{z}$. However, the evaluation of $\nabla f(\vec{z})$ is 
computationally expensive. It contains several matrix-matrix multiplications as
well as the evaluation of a matrix Jacobian and a Kronecker product. These
computations need to be done in every iteration. As we will see in
Lemma~\ref{lemma:def-p}, the evaluation of $\nabla f(\vec{z})$ can be done in a
way, so that the main effort lies in computing $n$ dyadics of vectors
$\vec{S}\left[-\vec{M}_j\vec{z},1\right]^\top\in\Real^{m,1}$ and
$\left(\vec{M}_j^\top\vec{M}_j\vec{z}\right)^\top\in\Real^{1,n}$.\par
In order to improve the performance of our numerical approach we further seek an
approximation to $\nabla f$ that requires significantly less floating point
operations. To this end, we assume for a moment that neither our matrix
$\vec{A}$, nor our vector $\vec{b}$ depend on the unknown $\vec{z}$. In that
case we obtain
\begin{equation}
      \label{eq:definition-q}
      \begin{split}
            \nabla f(\vec{z})&
            = \nabla \left( \frac{1}{2m}\lVert{}\vec{A}\vec{M}\vec{z}-\vec{b}\rVert{}^{2} \right)\\
            &=\frac{1}{m} (\vec{A}\vec{M})^{\top} (\vec{A}\vec{M}\vec{z} - \vec{b}) \eqqcolon \vec{q}
      \end{split}
\end{equation}
Our conclusions from~\eqref{eq:definition-q} are twofold. First of all,
$-\vec{q}$ seems to be a good candidate for a descent direction. At least when
our data $\vec{A}$ and $\vec{b}$ does not depend on $\vec{z}$, then $-\vec{q}$
is an optimal and significantly easier to evaluate descent direction. Secondly,
we can exploit~\eqref{eq:definition-q} to derive a refined version of the iPiano
algorithm for our task at hand. If we apply a lagged iteration on the descent
step of $f$, then our matrix $\vec{A}$ and our vector $\vec{b}$ become
automatically independent of our current iterate and $-\vec{q}$ would be the
steepest descent direction. The fact that $\vec{q}$ would not have to be
recomputed in every iteration could outweigh the loss of accuracy and yield 
an additional performance boost.\par
The following theorem states precise conditions under which the vector
$-\vec{q}$, defined in~\eqref{eq:definition-q}, yields a descent direction. Let
us emphasise that Theorem~\ref{thm:b_2} even allows a dependency on $\vec{z}$ in
$\vec{A}$ and $\vec{b}$.
\begin{theorem}
      \label{thm:b_2}
      The vector
      \begin{equation}
            -\vec{q}\coloneqq -\frac{1}{m}(\vec{A}(\vec{z})\vec{M})^{\top}
            (\vec{A}(\vec{z})\vec{M}\vec{z} - \vec{b}(\vec{z}))
      \end{equation}
      is a descent direction for $f(\vec{z})$ from~\eqref{eq:function_f}
      (resp.~\eqref{eq:a_grad_f}) at position $\vec{z}$ if the expression
      \begin{multline}
            \label{eq:b_9}
            \langle
            (\vec{A}(\vec{z})\vec{M}\vec{z}-\vec{b}(\vec{z})),\\
            \vec{A}(\vec{z})\vec{M} D[\vec{A}(\vec{z})\vec{M}\vec{z}-\vec{b}(\vec{z})]^{\top}
            (\vec{A}(\vec{z})\vec{M}\vec{z}-\vec{b}(\vec{z}))
            \rangle
      \end{multline}
      is non-negative. This is especially true if
      $\vec{A}(\vec{z})\vec{M}
      D[\vec{A}(\vec{z})\vec{M}\vec{z}-\vec{b}(\vec{z})]^{\top}$ is positive
      semi-definite.
      \begin{proof}
            Reordering the terms for $\nabla f(\vec{z})$ in~\eqref{eq:grad_f}
            yields the following relation between $\nabla f(\vec{z})$ and
            $\vec{q}$
            \begin{multline}
                  \label{eq:b_2}
                  \nabla f(\vec{z})\\ = \vec{q} +
                  \frac{1}{m}\bigl(
                  \underbrace{((\vec{M}\vec{z})^{\top}\otimes \vec{1})D[\vec{A}]-D[\vec{b}]}_{= D[\vec{A}\vec{M}\vec{z}-\vec{b}]-\vec{A}\vec{M}}
                  \bigr)^{\top}
                  (\vec{A}\vec{M}\vec{z}-\vec{b})
            \end{multline}
            where we have omitted the obvious dependencies on $\vec{z}$. Our
            vector $-\vec{q}$ will be a descent direction if
            $\langle -\vec{q}, \nabla f \rangle \leq 0$. Using~\eqref{eq:b_2}
            we conclude
            \begin{multline}
                  \label{eq:b_4}
                  \langle \vec{q}, \nabla f \rangle\\
                  = \langle \vec{q}, \vec{q}\rangle +
                  \frac{1}{m} \langle \vec{q}, (D[\vec{A}\vec{M}\vec{z}-\vec{b}]-\vec{A}\vec{M})^{\top}
                  (\vec{A}\vec{M}\vec{z}-\vec{b}) \rangle
            \end{multline}
            Expanding $\vec{q}$ in the second inner product yields
            \begin{equation}
                  \label{eq:b_5}
                  \begin{split}
                        &\langle \vec{q}, (D[\vec{A}\vec{M}\vec{z}-\vec{b}]-\vec{A}\vec{M})^{\top}
                        (\vec{A}\vec{M}\vec{z}-\vec{b}) \rangle \\
                        &= \langle
                        (\vec{A}\vec{M})^{\top}(\vec{A}\vec{M}\vec{z}-\vec{b}),\\
                        &\quad(D[\vec{A}\vec{M}\vec{z}-\vec{b}]-\vec{A}\vec{M})^{\top}(\vec{A}\vec{M}\vec{z}-\vec{b})
                        \rangle\\
                        &= \langle
                        (\vec{A}\vec{M})^{\top}(\vec{A}\vec{M}\vec{z}-\vec{b}), \\
                        &\quad D[\vec{A}\vec{M}\vec{z}-\vec{b}]^{\top} (\vec{A}\vec{M}\vec{z}-\vec{b})
                        \rangle
                        -
                        \langle \vec{q},\vec{q}\rangle
                  \end{split}
            \end{equation}
            Thus, we obtain
            \begin{equation}
                  \label{eq:b_6}
                  \begin{split}
                        &\langle \vec{q}, \nabla f \rangle\\
                        &=
                        \frac{1}{m}\langle
                        (\vec{A}\vec{M})^{\top}(\vec{A}\vec{M}\vec{z}-\vec{b}),
                        D[\vec{A}\vec{M}\vec{z}-\vec{b}]^{\top} (\vec{A}\vec{M}\vec{z}-\vec{b})
                        \rangle\\
                        &=
                        \frac{1}{m}\langle
                        (\vec{A}\vec{M}\vec{z}-\vec{b}), (\vec{A}\vec{M})
                        D[\vec{A}\vec{M}\vec{z}-\vec{b}]^{\top} (\vec{A}\vec{M}\vec{z}-\vec{b})
                        \rangle
                  \end{split}
            \end{equation}
            Now, we are in presence of a descent direction whenever the
            expression
            \begin{equation}
                  \label{eq:b_7}
                  \langle
                  (\vec{A}\vec{M}\vec{z}-\vec{b}), \vec{A}\vec{M}
                  D[\vec{A}\vec{M}\vec{z}-\vec{b}]^{\top} (\vec{A}\vec{M}\vec{z}-\vec{b})
                  \rangle
            \end{equation}
            is non-negative. This is especially true, if the matrix
            \begin{equation}
                  \label{eq:b_8}
                  \vec{A}(\vec{z}) \vec{M}
                  D[\vec{A}(\vec{z})\vec{M}\vec{z}-\vec{b}(\vec{z})]^{\top}
            \end{equation}
            is positive semi-definite.\qed
      \end{proof}
\end{theorem}
Let us conclude this section by remarking that the matrix
\begin{equation}
      \vec{A}(\vec{z}) \vec{M} D[\vec{A}(\vec{z})\vec{M}\vec{z}-\vec{b}(\vec{z})]^{\top}
\end{equation}
does not have any particular structure. Indeed, in general, it is made up from a
product of non-symmetric and non-square matrices. Thus, additional claims on the
spectral properties of this matrix are difficult to derive.\par
Nevertheless, we conjecture that
\begin{equation}
      \nabla f(\vec{z}) \approx \frac{1}{m} \left(\vec{A}(\vec{z})\vec{M} \right)^{\top}
      (\vec{A}(\vec{z})\vec{M}\vec{z} - \vec{b}(\vec{z}))
      \label{eq:grad_f_approx}
\end{equation}
is an efficient way to approximate $\nabla f(\vec{z})$ for computations. We will
investigate possible deficiencies later in our numerical experiments.
%
\subsection{Summary of the solution strategy}
\label{sec:an-inertial-proximal}
%
Our final algorithm for the computation of the depth and the albedo is given in
Algorithm~\ref{alg:iPianoPS}. For the step sizes we employed the ``lazy
backtracking'' algorithm as in~\cite{OCBP2013}. This includes increasing the
Lipschitz constant $L^{(\ell)}$ for $\nabla f$ by multiplication with a parameter $\eta>1$
($\eta=1.2$ in our experiments), until the new iterate $\tilde{\vec{z}}^{(l+1)}$
fulfils
\begin{multline}
      f(\tilde{\vec{z}}^{(\ell+1)})\leq f(\tilde{\vec{z}}^{(\ell)})\\
      +\langle \nabla f(\tilde{\vec{z}}^{l}),\tilde{\vec{z}}^{(\ell+1)}-\tilde{\vec{z}}^{(\ell)}\rangle
      +\frac{L^{(\ell)}}{2}\norm*{\tilde{\vec{z}}^{(\ell+1)}-\tilde{\vec{z}}^{(\ell)}}_2^2
      \label{eq:lazybacktr}
\end{multline}
The found Lipschitz constant $L^{(\ell)}$ divided by a $\mu\geq 1$ ($\mu=1.05$
in our experiments) delivers the start for estimating the Lipschitz constant
$L^{(\ell+1)}$ in the next iPiano iteration.\par
\begin{algorithm}
	\caption{Inertial Proximal Point Algorithm for Photometric Stereo}
	\label{alg:iPianoPS}
	\DontPrintSemicolon
	Choose prior $\vec{z}_0$ (classic \gls{PS}), prior weight $\lambda$
        ($10^{-6}$), $c>0$ ($0.01$) and $d>c$ ($1$)\;
	Initialise $\vec{z}^{(0)}$ ($\vec{z}_0$) and $\rho^{(0)}$ (classic
        \gls{PS}), and set $k=0$\;
	\Repeat{global convergence}{%
          Set $\tilde{\vec{z}}^{(0)} = \tilde{\vec{z}}^{(-1)} = \vec{z}^{(k)}$,
          $\delta^{(-1)}=d$ and $\ell=0$\;
          \Repeat{iPiano convergence}{
            Lipschitz constant $L^{(\ell)}$ estimation by lazy backtracking\;
            Aux. variable: $\nu=\dfrac{\delta^{(\ell-1)}+L^{(\ell)}/2}{c+L^{(\ell)}/2}$\;
            Step size updates: $\beta^{(\ell)}=\dfrac{\nu -1}{\nu+c-0.5}$ and
            $\alpha^{(\ell)} = \dfrac{1-\beta}{c+L^{(\ell)}/2}$\;
            Aux. variable: $\delta^{(\ell)}= \dfrac{1}{\alpha^{(\ell)}}-\dfrac{L^{(\ell)}}{2}-\dfrac{\beta^{(\ell)}}{\alpha^{(\ell)}}$\;
            Depth update: $\tilde{\vec{z}}^{(\ell+1)} = \prox_{\alpha^{(\ell)}g}
            \left( \tilde{\vec{z}}^{(\ell)} - \alpha^{(\ell)}
                  \nabla f\left( \tilde{\vec{z}}^{(\ell)} \right) +
                  \beta \left( \tilde{\vec{z}}^{(\ell)}-\tilde{\vec{z}}^{(\ell-1)}\right) \right)$\;
            $\ell = \ell+1$\;
          }
          $\vec{z}^{(k+1)} = \tilde{\vec{z}}^{(\ell+1)}$\;
          Albedo update using~\eqref{eq:rho_update}\;
          $k = k+1$\;
	}
\end{algorithm}
In Algorithm~\ref{alg:iPianoPS} we could also use a constant step size $\beta \in
[0,1[$, so that the computation of $\nu$ and $\delta^{(\ell)}$ would not be
required. By using $\beta=0.5$ in our numerical experiments we achieved
comparable results with respect to both computation time and quality of the
reconstructed surface. However, by applying a variable step size $\beta^{(\ell)}$
deduced from the proof of Lemma~4.6 in~\cite{OCBP2013} we ensure that the
auxiliary sequence $\lbrace \delta^{(\ell)}\rbrace_{\ell=-1}^{\infty}$ is
monotonically decreasing and therefore the convergence theory provided
in~\cite{OCBP2013} can be applied. To this end, let us remark that
$\norm*{\vec{z}}_2\rightarrow\infty$ implies
$g\left(\vec{z}\right)\rightarrow\infty$ and that $f$ is non-negative. Thus,
$f+g$ is coercive. Furthermore $g$ is convex and non-negative, such that $f+g$ is
bounded below. The function $f$ is obviously differentiable,
\cf~\eqref{eq:grad_f}.\par 
The final ingredient to apply the general convergence
result is the Lipschitz continuity of $\nabla f$, which we will investigate in
the following section. As a motivation, we recap the general convergence
result that was provided in~\cite{OCBP2013}, Theorem~4.8. For the definition of
Lipschitz continuity we refer to~\eqref{eq:def_Lipschitz}.
\begin{proposition}
      Let $\lbrace \tilde{\vec{z}}^{(\ell)}\rbrace_{\ell=0}^{\infty}$
      be a sequence generated by the inner loop of Algorithm~\ref{alg:iPianoPS}, with
      $\nabla f$ computed according to~\eqref{eq:grad_f}. If $\nabla f$ is
      Lipschitz continuous, then the following properties hold:
      \begin{enumerate}
      \item The sequence
            $\lbrace
                  f(\tilde{\vec{z}}^{(\ell)})+g(\tilde{\vec{z}}^{(\ell)})
            \rbrace_{\ell=0}^{\infty}$ converges.
      \item There exists a converging subsequence
            $\lbrace
                  \tilde{\vec{z}}^{(\ell_i)}\rbrace_{i=0}^{\infty}$.
      \item For any limit point
            $\tilde{\vec{z}}^*\coloneqq\lim_{i\rightarrow\infty}\tilde{\vec{z}}^{(\ell_i)}$
            we have
            \begin{equation}
                  0=\nabla f\left(\tilde{\vec{z}}^*\right)+\nabla g\left(\tilde{\vec{z}}^*\right)
            \end{equation}
            and
            \begin{equation}
                  \lim_{i\rightarrow\infty}f\left(\tilde{\vec{z}}^{(\ell_i)}\right)+g\left(\tilde{\vec{z}}^{(\ell_i)}\right)
                  =f\left(\tilde{\vec{z}}^*\right)+g\left(\tilde{\vec{z}}^*\right)
            \end{equation}
      \end{enumerate}
\end{proposition}
%
%
\section{Convergence analysis}
\label{sec:convAnalysis}
%
In Algorithm~\ref{alg:iPianoPS}, the Lipschitz constant $L$ of $\nabla
f(\tilde{\vec{z}})^{(\ell)}$ is estimated by a lazy backtracking strategy. To
derive a Lipschitz estimate for $\nabla f(\vec{z})$ for all $\vec{z}\in\Real^n$
and thereby ensure that the convergence theory for the iPiano algorithm provided
in~\cite{OCBP2013,P2016} can be applied, we first recall some general techniques
to combine Lipschitz estimates. Afterwards with these techniques we derive 
Lipschitz estimates for the gradient of $f$ as well as for our approximation of this 
gradient.\par
We conclude the convergence analysis by highlighting some aspects of the iPiano 
method. Thereby, we further justify our choice of a non-constant stepsize 
$\beta^{(\ell)}$, which might seem as a technical complication at first glance.
%
\subsection{Technical Preliminaries}
\label{sec:LipschitzTechniques}
%
Although the final Lipschitz estimates for $\nabla f$ and $\vec{q}$, that we are interested in, involves a
vector valued function with a vector valued input, to get there we will in 
the most general case discuss Lipschitz estimates for $\vec{F}:\Real^p\rightarrow\Real^{q,r}$ 
with different choices for $\vec{F}$, $p$, $q$ and $r$. The function $\vec{F}$ is Lipschitz 
continuous with a Lipschitz constant $L^{\vec{F}}$, if
\begin{equation}
      \norm*{\vec{F}(\vec{x})-\vec{F}(\vec{y})}_2\leq L^{\vec{F}}\norm*{\vec{x}-\vec{y}}_2
      \qquad\text{for all }\vec{x},\vec{y}\in\Real^{p}
      \label{eq:def_Lipschitz}
\end{equation}\par
The following lemma contains some basic techniques to combine Lipschitz
estimates. The proof is included for convenience.
\begin{lemma}
      \label{lemma:Lipschitz}
      Let $\vec{F}_1:\Real^m\rightarrow\Real^{n,p}$ and
      $\vec{F}_2:\Real^m\rightarrow\Real^{p,q}$ be Lipschitz continuous with
      $L^{(1)},L^{(2)}>0$, such that
      \begin{equation}
            \norm*{\vec{F}_k(\vec{x})-\vec{F}_k(\vec{y})}_2\leq L^{(k)}\norm*{\vec{x}-\vec{y}}_2,
            \label{eq:Lipschitz-functions}
      \end{equation}
      for all $\vec{x},\vec{y}\in\Real^m$ and
      $k\in\left\lbrace 1,2\right\rbrace$, then we have the following
      properties:
      \begin{enumerate}
      \item\label{prop:L1.i} If there exist $c_1,c_2$, such that
            $\norm*{\vec{F}_k(\vec{x})}_2\leq c_k$ for all $\vec{x}\in\Real^m$
            and $k\in\left\lbrace 1,2\right\rbrace$, then
            \begin{multline}
                  \norm*{\vec{F}_1(\vec{x})\vec{F}_2(\vec{x})-\vec{F}_1(\vec{y})\vec{F}_2(\vec{y})}_2\\
                  \leq \left(c_2L^{(1)}+c_1L^{(2)}\right)\norm*{\vec{x}-\vec{y}}_2
                  \label{eq:L1.i}
            \end{multline}
      \end{enumerate}
      The following properties only concern the scalar case.
      \begin{enumerate}[start=2]
      \item\label{prop:L1.ii} If $n=p=1$ and if there exists a $c_3>0$, such that
            $\vec{F}_1(\vec{x})\geq c_3$ for all $\vec{x}\in\Real^m$, then
            \begin{equation}
                  \abs*{\sqrt{\vec{F}_1(\vec{x})}-\sqrt{\vec{F}_1(\vec{y})}}\leq\frac{L^{(1)}}{2\sqrt{c_3}}\norm*{\vec{x}-\vec{y}}_2
                  \label{eq:L1.ii}
            \end{equation}
      \item\label{prop:L1.iii} If $n=p=1$ and if there exists a $c_4>0$, such that
            $\abs*{\vec{F}_1(\vec{x})}\geq c_4$ for all $\vec{x}\in\Real^m$,
            then
            \begin{equation}
                  \abs*{\frac{1}{\vec{F}_1(\vec{x})}-\frac{1}{\vec{F}_1(\vec{y})}}\leq\frac{L^{(1)}}{(c_4)^2}\norm*{ \vec{x}-\vec{y}}_2
                  \label{eq:L1.iii}
            \end{equation}
      \end{enumerate}
      \begin{proof}
            Let $\vec{x},\vec{y}\in\Real^m$.\par
            \begin{enumerate}
            \item Since
                  \begin{equation}
                        \begin{split}
                              &\norm*{\vec{F}_1(\vec{x})\vec{F}_2(\vec{x})-\vec{F}_1(\vec{y})\vec{F}_2(\vec{y})}_2\\
                              &= \left\lVert\vec{F}_1(\vec{x})\vec{F}_2(\vec{x})-\vec{F}_1(\vec{x})\vec{F}_2(\vec{y})\right.\\
                              &\quad\left.+\vec{F}_1(\vec{x})\vec{F}_2(\vec{y})-\vec{F}_1(\vec{y})\vec{F}_2(\vec{y})\right\rVert_2\\
                              & \leq \norm*{\vec{F}_1(\vec{x})}_2\norm*{\vec{F}_2(\vec{x})-\vec{F}_2(\vec{y})}_2\\
                              &\quad+\norm*{\vec{F}_2(\vec{y})}_2\norm*{\vec{F}_1(\vec{x})-\vec{F}_1(\vec{y})}_2\label{eq:L0}
                        \end{split}
                  \end{equation}
                  by $\norm*{\vec{F}_k(\vec{x})}_2\leq c_k$ for all $\vec{x}\in\Real^m$ and
                  $k\in\left\lbrace 1,2\right\rbrace$ and~\eqref{eq:Lipschitz-functions} we
                  get~\eqref{eq:L1.i}.
            \item Now let $n=p=1$.\par
                  If we have $\vec{F}_1(\vec{x})\geq c_3>0$ for all $\vec{x}\in\Real^m$, then
                  \begin{equation}
                        \begin{split}
                              &\abs*{\sqrt{\vec{F}_1(\vec{x})}-\sqrt{\vec{F}_1(\vec{y})}}
                              =\abs*{\frac{\vec{F}_1(\vec{x})-\vec{F}_1(\vec{y})}
                                {\sqrt{\vec{F}_1(\vec{x})}+\sqrt{\vec{F}_1(\vec{y})}}}\\
                              &\leq\abs*{\frac{1}{\sqrt{\vec{F}_1(\vec{x})}+\sqrt{\vec{F}_1(\vec{y})}}}
                              \abs*{\vec{F}_1(\vec{x})-\vec{F}_1(\vec{y})}
                        \end{split}
                  \end{equation}
                  and therefore, with~\eqref{eq:Lipschitz-functions} we get~\eqref{eq:L1.ii}.
            \item If we have $\abs*{\vec{F}_1(\vec{x})}\geq c_4>0$ for all $\vec{x}\in\Real^m$, then
                  \begin{equation}
                        \begin{split}
                              &\abs*{\frac{1}{\vec{F}_1(\vec{x})}-\frac{1}{\vec{F}_1(\vec{y})}}
                              =\abs*{\frac{\vec{F}_1(\vec{y})-\vec{F}_1(\vec{x})}{\vec{F}_1(\vec{x})\vec{F}_1(\vec{y})}}\\
                              &\leq\abs*{\frac{1}{\vec{F}_1(\vec{x})\vec{F}_1(\vec{y})}}\abs*{\vec{F}_1(\vec{x})-\vec{F}_1(\vec{y})}
                        \end{split}
                  \end{equation}
                  and with~\eqref{eq:Lipschitz-functions} we get~\eqref{eq:L1.iii}.\qed
            \end{enumerate}
      \end{proof}
\end{lemma}
%
\subsection{Lipschitz constant for the gradient of $f$}
\label{sec:Lipschitz-constants}
%
We investigate in this section the existence of a finite Lipschitz constant
$L^{\nabla f}$, such that for all $\vec{x}$, $\vec{y}\in\Real^n$
\begin{equation}
      \norm*{\nabla f(\vec{x})-\nabla f(\vec{y})}_2\leq
      L^{ \nabla f}\norm*{\vec{x}-\vec{y}}_2
\end{equation}
We will also investigate the existence of a Lipschitz constant $L^{\vec{q}}$ of
the approximated gradient $\vec{q}$ from~\eqref{eq:definition-q}, as well as the
dependencies of $L^{\nabla f}$ and $L^{\vec{q}}$ on $n$ and $m$.\par
We make the following assumptions:
\begin{description}
\item[(A1)] For all $j\in\left\lbrace 1,\dots,n\right\rbrace$ the approximation
      of the spatial gradient $\vec{M}_j\vec{z}$ is bounded, \ie~there is a
      $L^{\vec{z}}_j\in\left[0,\infty\right)$, such that $\norm*{ \vec{M}_j \vec{z}}_2
      \leq L^{\vec{z}}_j$ for all $\vec{z}\in\Real^n$.
\item[(A2)] For $n\rightarrow\infty$ we have $L^{\vec{z}}_j\rightarrow 0$ for
      all $j\in\left\lbrace 1,\dots,n\right\rbrace$, notably $L^{\vec{z}}_j\in
      O(1/\sqrt{n})$.
\end{description}
Let us remark that the previous assumptions on the decrease rate are done under
the assumption, that the grid step size of our image remains the same when the
number of pixels increases. While the finiteness of $L^{\nabla f}$ and
$L^{\vec{q}}$ hinges on~(A1), assumption~(A2) is only needed to derive the
dependencies of the Lipschitz constants on $n$. Although these are fairly strong
assumptions, we choose not to switch to a more restricted space than $\Real^n$,
and instead assume that the depth map that is to be reconstructed and also the
iterates $\tilde{\vec{z}}^{(\ell)}$ in Algorithm~\ref{alg:iPianoPS} fulfil~(A1)
and~(A2).\par
If (A1) holds true, then we additionally define
\begin{equation}
      \tilde{L}^{\vec{z}}_j\coloneqq \sqrt{1+\left(L^{\vec{z}}_j\right)^2}
      \qquad
      \text{for all }j\in\left\lbrace 1,\dots,n\right\rbrace
\end{equation}
so that we have
\begin{equation}
      \sqrt{1+\norm*{ \vec{M}_j \vec{z}}_2^2} \leq \tilde{L}^{\vec{z}}_j
      \qquad
      \text{for all }\vec{z}\in\Real^n
      \label{eq:L10.3}
\end{equation}
If (A2) holds true, then $\tilde{L}^{\vec{z}}_j\in O(1)$ for all
$j\in\left\lbrace 1,\dots,n\right\rbrace$.\par
To obtain a Lipschitz estimate of the gradient
\begin{multline}
      \label{eq:gradient-f}
      \nabla f(\vec{z}) = \frac{1}{m}
      \left( \left( (\vec{M}\vec{z})^{\top}\otimes \vec{1} \right) D[\vec{A}] +
            \vec{A}\vec{M} - D[\vec{b}]\right)^{\top}\\
      \left( \vec{A}\vec{M}\vec{z}-\vec{b} \right)
\end{multline}
we will first derive Lipschitz estimates for the individual components and then
combine them by using Lemma~\ref{lemma:Lipschitz}.
\begin{corollary}
      Let $j\in\left\lbrace 1,\dots,n\right\rbrace$ and
      $\vec{x},\vec{y}\in\Real^n$. In addition, we define $\kappa\coloneqq
      L^{\vec{z}}_j\norm*{\vec{M}_j}$. If (A1) holds true, then
      \begin{equation}
        \abs*{\norm*{\vec{M}_j\vec{x}}_2^2-\norm*{\vec{M}_j\vec{y}}_2^2}
          \leq 2\kappa\norm*{\vec{x}-\vec{y}}_2\label{eq:L10.5}
      \end{equation}
      \begin{equation}
         \abs*{\sqrt{1+\norm*{\vec{M}_j\vec{x}}_2^2}-\sqrt{1+\norm*{\vec{M}_j\vec{y}}_2^2}}
         \leq \kappa\norm*{\vec{x}-\vec{y}}_2\label{eq:L11}
      \end{equation}
      \begin{equation}
        \abs*{\frac{1}{\sqrt{1+\norm*{\vec{M}_j\vec{x}}_2^2}}-\frac{1}{\sqrt{1+\norm*{\vec{M}_j\vec{y}}_2^2}}}
         \leq \kappa\norm*{\vec{x}-\vec{y}}_2\label{eq:L12}
      \end{equation}
      \begin{equation}
        \abs*{\frac{1}{1+\norm*{\vec{M}_j\vec{x}}_2^2}-\frac{1}{1+\norm*{\vec{M}_j\vec{y}}_2^2}}
        \leq 2\kappa\norm*{\vec{x}-\vec{y}}_2\label{eq:L12.5}
      \end{equation}
      \begin{equation}
            \abs*{\frac{1}{\sqrt{1+\norm*{\vec{M}_j\vec{x}}_2^2}^3}-\frac{1}{\sqrt{1+\norm*{\vec{M}_j\vec{y}}_2^2}^3}}
            \leq 3\kappa\norm*{\vec{x}-\vec{y}}_2\label{eq:L13}
      \end{equation}
      \begin{align}
        &\norm*{
          \begin{bmatrix}-\vec{M}_j\vec{x}\\1\end{bmatrix}\left(\vec{M}_j\vec{x}\right)^\top
        -\begin{bmatrix}-\vec{M}_j\vec{y}\\1\end{bmatrix}\left(\vec{M}_j\vec{y}\right)^\top
        }_2 \nonumber\\
        &\qquad\leq \left(\tilde{L}^{\vec{z}}_j+L^{\vec{z}}_j\right)\norm*{\vec{M}_j}\norm*{\vec{x}-\vec{y}}_2\label{eq:L13.5}
      \end{align}
      \begin{proof}
            Using Lemma~\ref{lemma:Lipschitz}.\ref{prop:L1.i} with
            $\vec{F}_1(\vec{z})\coloneqq\left(\vec{M}_j\vec{z}\right)^\top$,
            $\vec{F}_2(\vec{z})\coloneqq\vec{M}_j\vec{z}$, (A1) and
            $\norm*{\vec{M}_j\vec{x}-\vec{M}_j\vec{y}}_2\leq\norm*{\vec{M}_j}\norm*{\vec{x}-\vec{y}}_2$
            we can deduce~\eqref{eq:L10.5}.\par
             Lemma~\ref{lemma:Lipschitz}.\ref{prop:L1.ii},
            $\vec{F}_1(\vec{z})\coloneqq1+\norm*{\vec{M}_j\vec{z}}_2^2\geq 1$
            for all $\vec{z}\in\Real^n$ and the just shown validity
            of~\eqref{eq:L10.5} we get~\eqref{eq:L11}.\par
            Making use of Lemma~\ref{lemma:Lipschitz}.\ref{prop:L1.iii},
            $\vec{F}_1(\vec{z})\coloneqq\sqrt{1+\norm*{\vec{M}_j\vec{z}}_2^2}\geq
            1$ holds for all $\vec{z}\in\Real^n$ and by employing~\eqref{eq:L11}
            we obtain~\eqref{eq:L12}.\par
            By using Lemma~\ref{lemma:Lipschitz}.\ref{prop:L1.iii},
            $\vec{F}_1(\vec{z})\coloneqq 1+\norm*{\vec{M}_j\vec{z}}_2^2\geq 1$
            for all $\vec{z}\in\Real^n$, and together with~\eqref{eq:L10.5} we
            get~\eqref{eq:L12.5}.\par
            By using Lemma~\ref{lemma:Lipschitz}.\ref{prop:L1.i},
            $\vec{F}_1(\vec{z})\coloneqq
            1/\sqrt{1+\norm*{\vec{M}_j\vec{z}}_2^2}\leq 1$ and
            $\vec{F}_2(\vec{z})\coloneqq
            1/\left(1+\norm*{\vec{M}_j\vec{z}}_2^2\right)\leq 1$ hold for all
            $\vec{z}\in\Real^n$, so that together with~\eqref{eq:L12}
            and~\eqref{eq:L12.5} it is easy to see that~\eqref{eq:L13} is
            true.\par
            By combining Lemma~\ref{lemma:Lipschitz}.\ref{prop:L1.i},
            $\vec{F}_1(\vec{z})\coloneqq\left[-\vec{M}_j\vec{z},1\right]^\top$,
            $\vec{F}_2(\vec{z})\coloneqq\left(\vec{M}_j\vec{z}\right)^\top$,
            (A1),~\eqref{eq:L10.3} and
            $\norm*{\vec{M}_j\vec{x}-\vec{M}_j\vec{y}}_2\leq\norm*{\vec{M}_j}\norm*{\vec{x}-\vec{y}}_2$
            we finally get the validity of~\eqref{eq:L13.5}.
            \qed
      \end{proof}
\end{corollary}
The following lemma contains the first indication of Lipschitz estimates.
\begin{lemma}
      Let $\vec{x},\vec{y}\in\Real^n$ and $\vec{A}$ be defined as
      in~\eqref{eq:definitionA}. If (A1) holds true, then
      \begin{equation}
            \norm*{\vec{A}_j(\vec{x})-\vec{A}_j(\vec{y})}_2
            \leq \underbrace{\rho_j\norm*{\vec{S}_{\ell}}_2L^{\vec{z}}_j\norm*{\vec{M}_j}}_{\eqqcolon L^{\vec{A}}_j}
            \norm{\vec{x}-\vec{y}}_2
            \label{eq:L14}
      \end{equation}
      for all $j\in\left\lbrace 1,\dots,n\right\rbrace$ as well as
      \begin{equation}
            \norm*{\vec{A}(\vec{x})-\vec{A}(\vec{y})}_2
            \leq\Big(\underbrace{\max_jL^{\vec{A}}_j}_{\eqqcolon L^{\vec{A}}}\Big)\norm*{\vec{x}-\vec{y}}_2
            \label{eq:L15.13}
      \end{equation}
      If additionally (A2) holds true, then
      \begin{align}
        L^{\vec{A}}_j & \in O(\sqrt{m/n})\qquad\text{for all }j\in\left\lbrace 1,\dots,n\right\rbrace
                        \label{eq:L15.14}\\
        L^{\vec{A}} & \in O(\sqrt{m/n})
                      \label{eq:L15.15}
      \end{align}
      \begin{proof}
            Let $j\in\left\lbrace 1,\dots,n\right\rbrace$. From the definition
            of $\vec{A}_j$ in~\eqref{eq:18} and from~\eqref{eq:L12} follows
            directly~\eqref{eq:L14}.\par
            By $\rho_j\in\left[0,1\right]$,
            $\norm*{\vec{S}_{\ell}}_2\in O(\sqrt{m})$, (A2) and
            $\norm*{\vec{M}_j}\in O(1)$ we obtain~\eqref{eq:L15.14}.\par
            Since $\vec{A}$ is a (in general non-square) block diagonal matrix
            we have
            \begin{equation}
                  \vec{A}^\top \vec{A}
                  =\begin{bmatrix}
                        \vec{A}_1^\top \vec{A}_1 & & \\
                        & \ddots &\\
                        & & \vec{A}_n^\top \vec{A}_n
                  \end{bmatrix}\in\Real^{2n,2n},\label{eq:L15.10}
            \end{equation}
            and with
            \begin{multline}
                  \det\left(\vec{A}^\top \vec{A}-\lambda \vec{1}_{2n}\right)\\
                  =\det\left(\vec{A}_1^\top \vec{A}_1-\lambda \vec{1}_2\right) \dots
                  \det\left(\vec{A}_n^\top \vec{A}_n-\lambda \vec{1}_2\right)\label{eq:L15.11}
            \end{multline}
            for all $\lambda\in\Real$ we have
            \begin{equation}
                  \begin{split}
                        &\norm*{\vec{A}}_2\\
                        &= \sqrt{\max\left\lbrace\mathrm{eig}\left(\vec{A}^\top
                                      \vec{A}\right)\right\rbrace}\\
                        &= \sqrt{\max\left\lbrace\mathrm{eig}\left(\vec{A}_1^\top
                                      \vec{A}_1\right),\dots,\mathrm{eig}\left(\vec{A}_n^\top
                                      \vec{A}_n\right)\right\rbrace}\\
                        &= \max_j\sqrt{\max\left\lbrace\mathrm{eig}\left(\vec{A}_j^\top
                                      \vec{A}_j\right)\right\rbrace}=\max_j\norm*{\vec{A}_j}_2.\label{eq:L15.12}
                  \end{split}
            \end{equation}
            In the same way we can derive the equation
            \begin{equation}
                  \norm*{\vec{A}(\vec{x})-\vec{A}(\vec{y})}_2 =
                  \max_j\norm*{\vec{A}_j(\vec{x})-\vec{A}_j(\vec{y})}_2
            \end{equation}
            and with~\eqref{eq:L14} we obtain~\eqref{eq:L15.13}.\par
            From~\eqref{eq:L15.13} and~\eqref{eq:L15.14} follows~\eqref{eq:L15.15}.
            \qed
      \end{proof}
\end{lemma}
The following assertion is an immediate consequence of~\eqref{eq:L15.13}.
\begin{corollary}
      Let $\vec{x},\vec{y}\in\Real^n$. If (A1) holds true, then
      \begin{equation}
            \norm*{\vec{A}(\vec{x})\vec{M}-\vec{A}(\vec{y})\vec{M}}_2
            \leq L^{\vec{A}}\norm*{\vec{M}}_2\norm*{\vec{x}-\vec{y}}_2\label{eq:L16}
      \end{equation}
\end{corollary}
We proceed with another building block used for coming to 
Proposition~\ref{prop:LipschitzConstants}.
\begin{corollary}
      Let $\vec{x},\vec{y}\in\Real^n$. If (A1) holds true, then
      \begin{multline}
                  \norm*{\vec{A}_j(\vec{x})\vec{M}_j\vec{x}-\vec{b}_j(\vec{x})-\vec{A}_j(\vec{y})\vec{M}_j\vec{y}+\vec{b}_j(\vec{y})}_2 \\
                  \leq \underbrace{\rho_j\norm*{\vec{S}}_2\norm*{\vec{M}_j}\left(\tilde{L}^{\vec{z}}_jL^{\vec{z}}_j+1\right)}_{\eqqcolon L^f_j}\norm*{\vec{x}-\vec{y}}_2
               \label{eq:L15.3}
      \end{multline}
      for all $j\in\left\lbrace 1,\dots,n\right\rbrace$ and
      \begin{multline}
            \norm*{\vec{A}(\vec{x})\vec{M}\vec{x}-\vec{b}(\vec{x})-\vec{A}(\vec{y})\vec{M}\vec{y}+\vec{b}(\vec{y})}_2\\
            \leq\sqrt{\sum_{j=1}^{n}\left(L^f_j\right)^2}\norm*{\vec{x}-\vec{y}}_2
            \eqqcolon L^f\norm*{\vec{x}-\vec{y}}_2\label{eq:L15.4}
      \end{multline}
      If additionally (A2) holds true, then for all $j\in\left\lbrace 1,\dots,n\right\rbrace$
      \begin{equation}
            L^f_j\in O(\sqrt{m}),\qquad L^f\in O(\sqrt{mn})\label{eq:L15.45}
      \end{equation}
      \begin{proof}
            From Lemma~\ref{lemma:Lipschitz}.\ref{prop:L1.i} with
            \begin{equation}
                  \vec{F}_1(\vec{z})\coloneqq \frac{\rho_j}{\sqrt{1+\norm*{\vec{M}_j\vec{z}}}}\vec{S},\qquad
                  \vec{F}_2(\vec{z})\coloneqq
                  \begin{bmatrix}
                        -\vec{M}_j\vec{z} \\ 1
                  \end{bmatrix}
            \end{equation}
            $\norm*{\vec{F}_1(\vec{z})}_2\leq \rho_j\norm*{\vec{S}}_2$,~\eqref{eq:L10.3},~\eqref{eq:L12} and
            the Lipschitz estimate
            $\norm*{\vec{F}_2(\vec{x})-\vec{F}_2(\vec{x})}_2\leq\norm*{\vec{M}_j}\norm*{\vec{x}-\vec{x}}_2$
            follows~\eqref{eq:L15.3}.\par
            From~\eqref{eq:L15.3} and with the definition of the Euclidean norm
            we obtain~\eqref{eq:L15.4}.\par
            The inclusion~\eqref{eq:L15.45} follows from (A2), $\rho_j\in O(1)$ and $\norm*{\vec{M}_j}\in O(1)$ for all
            $j\in\left\lbrace 1,\dots,n\right\rbrace$ and $\norm*{\vec{S}}_2\in O(\sqrt{m})$.
            \qed
      \end{proof}
\end{corollary}
The following lemma will subsequently be used to derive a Lipschitz estimate for
$\nabla f$, but it also shows a more explicit representation of the exact
gradient.
\begin{lemma}
      \label{lemma:def-p}
      For $\vec{z}\in\Real^n$ we have
      \begin{equation}
            \begin{split}
                  \vec{p}(\vec{z}) & \coloneqq
                  \left(\left(\vec{M}\vec{z}\right)^\top\otimes\vec{1}_{mn}\right)D\left[\vec{A}(\vec{z})\right](\vec{z})-D\left[\vec{b}(\vec{z})\right](\vec{z})\\
                  & =\begin{bmatrix}
                        -\frac{\rho_1}{\sqrt{1+\norm*{\vec{M}_1\vec{z}}_2^2}^3}\vec{S}\begin{bmatrix}-\vec{M}_1\vec{z}\\1\end{bmatrix}\left(\vec{M}_1^\top\vec{M}_1\vec{z}\right)\top\\
                        \vdots\\
                        -\frac{\rho_n}{\sqrt{1+\norm*{\vec{M}_n\vec{z}}_2^2}^3}\vec{S}\begin{bmatrix}-\vec{M}_1\vec{z}\\1\end{bmatrix}\left(\vec{M}_n^\top\vec{M}_n\vec{z}\right)^\top
                  \end{bmatrix}
                  \label{eq:L15.48}
            \end{split}
      \end{equation}
      \begin{proof}
            To find an expression for $\vec{p}$ without the Kronecker product,
            we will simply write down all components of $\vec{p}$ and
            consecutively join them.\par
            For $\vec{z}\in\Real^n$ and $j\in\left\lbrace 1,\dots,n\right\rbrace$ we have
            \begin{align}
              \vec{b}_j= &\, \vec{I}_j-\frac{\rho_j}{\sqrt{1+\norm*{\vec{M}_j\vec{z}}_2^2}}\vec{S}_r,\\
              D\left[ \vec{b}_j\right]= &\, \frac{\rho_j}{\sqrt{1+\norm*{\vec{M}_j\vec{z}}_2^2}^3}\vec{S}_r \left(\vec{M}_j^\top \vec{M}_j \vec{z}\right)^\top\in\Real^{m,n}
            \end{align}
            leading to
            \begin{equation}
                  D\left[\vec{b}(\vec{z})\right](\vec{z})=
                  \begin{bmatrix}
                        \frac{\rho_1}{\sqrt{1+\norm*{\vec{M}_1\vec{z}}_2^2}^3}\vec{S}_r\left(\vec{M}_1^\top\vec{M}_1\vec{z}\right)^\top\\
                        \vdots\\
                        \frac{\rho_n}{\sqrt{1+\norm*{\vec{M}_n\vec{z}}_2^2}^3}\vec{S}_r\left(\vec{M}_n^\top\vec{M}_n\vec{z}\right)^\top
                  \end{bmatrix}
                  \in\Real^{mn,n}
                  \label{eq:L15.5}
            \end{equation}
            For $k\in\left\lbrace 1,2\right\rbrace$ with
            $\vec{A}^{k}_j(\vec{z})$, $\vec{S}^{k}_{\ell}$ and $\vec{M}_j^k$ we
            denote the $k$-th column of $\vec{A}_j(\vec{z})$ and $\vec{S}_{\ell}$
            and the $k$-th row of $\vec{M}_j$. We derive
            \begin{equation}
                  D\left[\vec{A}_j^{k}(\vec{z})\right](\vec{z})=\frac{\rho_j \vec{S}_{\ell}^{k}}{\sqrt{1+\norm*{\vec{M}_j\vec{z}}_2^2}^3}\left(\vec{M}_j^\top \vec{M}_j \vec{z}\right)^\top\in\Real^{m,n}
            \end{equation}
            Because of the structure of $\vec{A}$, defined in~\eqref{eq:definitionA}, 
            and our choice of the Jacobian matrix as per
            Definition~\ref{thm:MatrixJacobian}, the Jacobian matrix of
            $\vec{A}$ has the form
            \begin{equation}
                  D\left[\vec{A}(\vec{z})\right](\vec{z})=
                  \begin{bmatrix}
                        D\left[\vec{A}_1^{1}(\vec{z})\right](\vec{z})\\
                        \vec{0}_{mn-m,n}\\
                        D\left[\vec{A}_1^{2}(\vec{z})\right](\vec{z})\\
                        \vec{0}_{mn,n}\\
                        D\left[\vec{A}_2^{1}(\vec{z})\right](\vec{z})\\
                        \vec{0}_{mn-m,n}\\
                        D\left[\vec{A}_2^{2}(\vec{z})\right](\vec{z})\\
                        \vec{0}_{mn,n}\\
                        \vdots\\
                        D\left[\vec{A}_n^{1}(\vec{z})\right](\vec{z})\\
                        \vec{0}_{mn-m,n}\\
                        D\left[\vec{A}_n^{2}(\vec{z})\right](\vec{z})
                  \end{bmatrix}
                  \in\Real^{2mn^2,n}
            \end{equation}
            where $\vec{0}_{p,q}$ is a $p\times q$ block of zeros. Since
            \begin{multline}
                  \Real^{mn,2mn^2}\ni\left(\vec{M}\vec{z}\right)^\top\otimes\vec{1}_{mn}\\=
                  \begin{bmatrix}
                        \vec{M}_1^1\vec{z}& &  & \vec{M}_1^2\vec{z}& &  &        & \vec{M}_n^2\vec{z}& &  \\
                        & \ddots &             & & \ddots &             & \cdots & &  \ddots &            \\
                        & & \vec{M}_1^1\vec{z} & & & \vec{M}_1^2\vec{z} &        & & & \vec{M}_n^2\vec{z}
                  \end{bmatrix}
            \end{multline}
            we get
            \begin{equation}
                  \begin{split}
                        &\left(\left(\vec{M}\vec{z}\right)^\top\otimes\vec{1}_{mn}\right)D\left[\vec{A}(\vec{z})\right](\vec{z})
                        \\&=
                        \begin{bmatrix}
                              \vec{M}_1^1D\left[\vec{A}_1^{1}(\vec{z})\right](\vec{z})
                              +\vec{M}_1^2\vec{z}D\left[\vec{A}_1^{2}(\vec{z})\right](\vec{z})\\
                              \vdots\\
                              \vec{M}_n^1\vec{z}D\left[\vec{A}_n^{1}(\vec{z})\right](\vec{z})
                              +\vec{M}_n^2\vec{z}D\left[\vec{A}_n^{2}(\vec{z})\right](\vec{z})
                        \end{bmatrix}
                        \\&=
                        \begin{bmatrix}
                              \frac{\rho_1}{\sqrt{1+\norm*{\vec{M}_1\vec{z}}_2^2}^3}\vec{S}_\ell\vec{M}_1\vec{z}\left(\vec{M}_1^\top\vec{M}_1\vec{z}\right)\top\\
                              \vdots\\
                              \frac{\rho_n}{\sqrt{1+\norm*{\vec{M}_n\vec{z}}_2^2}^3}\vec{S}_\ell\vec{M}_n\vec{z}\left(\vec{M}_n^\top\vec{M}_n\vec{z}\right)^\top
                        \end{bmatrix}
                  \end{split}
            \end{equation}
            Therefore, together with~\eqref{eq:L15.5}, we get
            \begin{multline}
                  \vec{p}(\vec{z})=
                  \begin{bmatrix}
                        -\frac{\rho_1}{\sqrt{1+\norm*{\vec{M}_1\vec{z}}_2^2}^3}\vec{S}\begin{bmatrix}-\vec{M}_1\vec{z}\\1\end{bmatrix}\left(\vec{M}_1^\top\vec{M}_1\vec{z}\right)\top\\
                        \vdots\\
                        -\frac{\rho_n}{\sqrt{1+\norm*{\vec{M}_n\vec{z}}_2^2}^3}\vec{S}\begin{bmatrix}-\vec{M}_1\vec{z}\\1\end{bmatrix}\left(\vec{M}_n^\top\vec{M}_n\vec{z}\right)^\top
                  \end{bmatrix}
            \end{multline}
            \qed
      \end{proof}
\end{lemma}
\begin{corollary}
      Let $\vec{z}\in\Real^n$, $j\in\left\lbrace1,\dots,n\right\rbrace$ and
      \begin{equation}
            \vec{p}_j(\vec{z})\coloneqq
            -\frac{\rho_j}{\sqrt{1+\norm*{\vec{M}_j\vec{z}}_2^2}^3}\vec{S}
            \begin{bmatrix}-\vec{M}_j\vec{z}\\1\end{bmatrix}\left(\vec{M}_j^\top\vec{M}_j\vec{z}\right)^\top
      \end{equation}
      If (A1) holds true, then
      \begin{multline}
                  \norm*{\vec{p}_j(\vec{x})-\vec{p}_j(\vec{y})}_2\\
                  \leq \underbrace{\rho_j\norm*{\vec{S}}_2\norm*{\vec{M}_j}^2\left(3\tilde{L}^{\vec{z}}_j\left(L^{\vec{z}}_j\right)^2+\tilde{L}^{\vec{z}}_j+L^{\vec{z}}_j\right)}_{\eqqcolon L^{\vec{p}}_j}
                  \norm*{\vec{x}-\vec{y}}_2\label{eq:L20}
      \end{multline}
      If additionally (A2) holds true, then
      \begin{equation}
            L^{\vec{p}}_j\in O(\sqrt{m})\label{eq:L21}
      \end{equation}
      \begin{proof}
            With Lemma~\ref{lemma:Lipschitz}.\ref{prop:L1.i} and the settings
            \begin{align}
              \vec{F}_1(\vec{z}) & \coloneqq -\frac{\rho_j}{\sqrt{1+\norm*{\vec{M}_j\vec{z}}_2}^3}\vec{S}\\
              \vec{F}_2(\vec{z}) & \coloneqq \left[-\vec{M}_j\vec{z},1\right]^\top\left(\vec{M}_j\vec{z}\right)^\top\vec{M}_j
            \end{align}
            and with~\eqref{eq:L13},~\eqref{eq:L13.5} and (A1) we obtain~\eqref{eq:L20}.\par
            Equation~\eqref{eq:L21} follows from
            $\norm*{\vec{S}}_2\in O(\sqrt{m})$ and the estimate
            $\left(3\tilde{L}^{\vec{z}}_j\left(L^{\vec{z}}_j\right)^2+\tilde{L}^{\vec{z}}_j+L^{\vec{z}}_j\right)\in
            O(1)$ according to (A2).
            \qed
      \end{proof}
\end{corollary}
Let us now present finally the main result of this section.
\begin{proposition}
      \label{prop:LipschitzConstants}
      Let $\nabla f$ be defined as in~\eqref{eq:grad_f} and $\vec{q}$ be defined as
      in~\eqref{eq:definition-q}, $\vec{x},\vec{y}\in\Real^n$. If (A1) holds true,
      then
      \begin{align}
        \norm*{\nabla f(\vec{x})-\nabla f(\vec{y})}_2 & \leq L^{\nabla f}\norm*{\vec{x}-\vec{y}}_2\label{eq:L24}\\
        \norm*{\vec{q}(\vec{x})-\vec{q}(\vec{y})}_2 & \leq L^{\vec{q}}\norm*{\vec{x}-\vec{y}}_2\label{eq:L25}
      \end{align}
      where
      \begin{align}
        L^{\nabla f} & \coloneqq
                       \frac{1}{m}\Bigg(\sqrt{\sum_{j=1}^n\left(\norm*{\vec{I}_j}^2_2+\rho_j^2\norm*{\vec{S}}^2_2\right)}\nonumber\\
                     &\quad\left(\sqrt{\sum_{j=1}^n\left(L^{\vec{p}}_j\right)^2}+L^{\vec{A}}\norm*{\vec{M}}_2\right)\nonumber\\
                     &\quad+L^f\Bigg(\max_j \rho_j\norm*{\vec{S}_{\ell}}_2\norm*{\vec{M}}_2\nonumber\\
                     &\quad+\sqrt{\sum_{j=1}^n \rho_j^2\norm*{\vec{S}}^2_2\left(\tilde{L}^{\vec{z}}_j L^{\vec{z}}_j\right)^2\norm*{\vec{M}_j}^2_2}\Bigg)\Bigg)\\
        L^{\vec{q}} & \coloneqq
                      \frac{1}{m}\Bigg(\sqrt{\sum_{j=1}^{n}\left(\norm*{\vec{I}_j}^2_2+\rho_j^2\norm*{\vec{S}}^2_2\right)}L^{\vec{A}}\norm*{\vec{M}}_2\nonumber\\
                     &\quad +\max_j\rho_{j}\norm*{\vec{S}_{\ell}}_2\norm*{\vec{M}}_2L^f\Bigg)
      \end{align}
      If additionally (A2) holds true, then
      \begin{align}
        L^{\nabla f} & \in O(n)\label{eq:L28}\\
        L^{\vec{q}} & \in O(\sqrt{n})\label{eq:L29}
      \end{align}
      \begin{proof}
            First we will derive a Lipschitz estimate for $\vec{q}$. Assume that (A1) holds true. We define
            \begin{align}
              \vec{F}_1(\vec{z}) & \coloneqq \frac{1}{m} \left(\vec{A}(\vec{z})\vec{M}\right)^{\top} \\
              \vec{F}_2(\vec{z}) & \coloneqq \vec{A}(\vec{z})\vec{M}\vec{z} - \vec{b}(\vec{z})
            \end{align}
            for all $\vec{z}\in\Real^n$. As in~\eqref{eq:L15.12} we get
            \begin{equation}
                  \norm*{\vec{F}_1(\vec{z})}_2\leq \frac{1}{m}\max_j \rho_j\norm*{\vec{S}_\ell}_2\norm*{\vec{M}}_2
                  \label{eq:L15.39}
            \end{equation}
            and we also have
            \begin{equation}
                  \norm*{\vec{F}_2(\vec{z})}_2\leq \sqrt{\sum_{j=1}^{n}\left(\norm*{\vec{I}_j}^2_2+\rho_j^2\norm*{\vec{S}}_2^2\right)}
                  \label{eq:L15.391}
            \end{equation}
            With Lemma~\ref{lemma:Lipschitz}.\ref{prop:L1.i} and the Lipschitz estimates~\eqref{eq:L16} and~\eqref{eq:L15.4}
            we get~\eqref{eq:L25}.\par
            To deduce the Lipschitz estimate for $\nabla f$, we extend the proof by redefining
            \begin{equation}
                  \vec{F}_1(\vec{z}) \coloneqq \frac{1}{m} \left(\vec{A}(\vec{z})\vec{M}+\vec{p}(\vec{z})\right)^{\top}
            \end{equation}
            where $\vec{p}$ is defined as in~\eqref{eq:L15.48}. By~\eqref{eq:L20} we get
            \begin{equation}
                  \norm*{\vec{p}(\vec{x})-\vec{p}(\vec{y})}_2\leq \sqrt{\sum_{j=1}^n \left(L^{\vec{p}}_j\right)^2}\norm*{\vec{x}-\vec{y}}_2
            \end{equation}
            Together with~\eqref{eq:L16} we obtain
            \begin{multline}
                  \norm*{\vec{F}_1(\vec{x})-\vec{F}_1(\vec{y})}_2\\
                  \leq\frac{1}{m}\left(L^{\vec{A}}\norm*{\vec{M}}_2+\sqrt{\sum_{j=1}^n\left(L^{\vec{p}}_j\right)^2}\right)
                  \norm*{\vec{x}-\vec{y}}_2
                  \label{eq:L22}
            \end{multline}
            Furthermore by the definition of $\vec{p}$ in~\eqref{eq:L15.48} and analogously to~\eqref{eq:L15.39} we get
            \begin{multline}
                  \norm*{\vec{F}_1(\vec{z})}_2\leq \frac{1}{m}\Bigg(\max_j \rho_j\norm*{\vec{S}_{\ell}}_2\norm*{\vec{M}}_2\\
                  +\sqrt{\sum_{j=1}^n \rho_j^2\norm*{\vec{S}}^2_2\left(\tilde{L}^{\vec{z}}_j L^{\vec{z}}_j\right)^2\norm*{\vec{M}_j}^2_2}\Bigg)
                  \label{eq:L23}
            \end{multline}
            Now by Lemma~\ref{lemma:Lipschitz}.\ref{prop:L1.i} with~\eqref{eq:L15.4},~\eqref{eq:L15.391},~\eqref{eq:L22}
            and~\eqref{eq:L23} we can deduce~\eqref{eq:L24}.\par
            Now assume that (A2) holds true. The inclusions~\eqref{eq:L15.15}, $\rho_j\in O(1)$, $\norm*{\vec{M}}_2\in O(1)$, $\norm*{\vec{I}_j}_2\in O(\sqrt{m})$ and
            $\norm*{\vec{S}}_2\in O(\sqrt{m})$ lead to
            \begin{equation}
                  \sqrt{\sum_{j=1}^n\left(\norm*{\vec{I}_j}^2_2+\rho_j^2\norm*{\vec{S}}^2_2\right)}L^{\vec{A}}\norm*{\vec{M}}_2\in O(m)
                  \label{eq:L30}
            \end{equation}
            Furthermore by~\eqref{eq:L15.45}, $\rho_j\in O(1)$, $\norm*{\vec{M}}_2\in O(1)$ and $\norm*{\vec{S}_\ell}_2\in O(\sqrt{m})$ we obtain
            \begin{equation}
                  \max_j \rho_j\norm*{\vec{S}_\ell}_2\norm*{\vec{M}}_2 L^f\in O(m\sqrt{n})
                  \label{eq:L31}
            \end{equation}
            Now from~\eqref{eq:L30},~\eqref{eq:L31} and $1/m\in O(1/m)$ follows~\eqref{eq:L29}.\par
            The inclusions~\eqref{eq:L21}, $\rho_j\in O(1)$, $\norm*{\vec{M}}_2\in O(1)$, $\norm*{\vec{I}_j}_2\in O(\sqrt{m})$ and
            $\norm*{\vec{S}}_2\in O(\sqrt{m})$ lead to
            \begin{equation}
                  \sqrt{\sum_{j=1}^n\left(\norm*{\vec{I}_j}^2_2+\rho_j^2\norm*{\vec{S}}^2_2\right)}\sqrt{\sum_{j=1}^n\left(L^{\vec{p}}_j\right)^2}\in O(mn)
                  \label{eq:L32}
            \end{equation}
            By~\eqref{eq:L15.45}, $\rho_j\in O(1)$, $\norm*{\vec{M}_j}\in O(1)$, $\tilde{L}^{\vec{z}}_j\in O(1)$,
            $L^{\vec{z}}_j\in O(1/\sqrt{n})$ and $\norm*{\vec{S}}_2\in O(\sqrt{m})$ we obtain
            \begin{equation}
                  L^f\sqrt{\sum_{j=1}^n\rho_j^2\norm*{\vec{S}}_2^2\left(\tilde{L}^{\vec{z}}_j L^{\vec{z}}_j\right)^2\norm*{\vec{M}_j}^2}\in O(m)
                  \label{eq:L33}
            \end{equation}
            Finally from~\eqref{eq:L30},~\eqref{eq:L31},~\eqref{eq:L32},~\eqref{eq:L33} and $1/m\in O(1/m)$ follows~\eqref{eq:L28}.
            \qed
      \end{proof}
\end{proposition}
We have shown that under the assumptions (A1) and (A2) the gradient as well as the
approximated gradient of $f$ are Lipschitz continuous.\par
As already indicated in~\eqref{eq:lazybacktr}, for practical applications we may
also be interested in local Lipschitz constants $L^{(\ell)}$ fulfilling
\begin{multline}
      f(\tilde{\vec{z}}^{(\ell+1)})\leq f(\tilde{\vec{z}}^{(\ell)})\\
      +\langle \nabla f(\tilde{\vec{z}}^{l}),\tilde{\vec{z}}^{(\ell+1)}-\tilde{\vec{z}}^{(\ell)}\rangle
      +\frac{L^{(\ell)}}{2}\norm*{\tilde{\vec{z}}^{(\ell+1)}-\tilde{\vec{z}}^{(\ell)}}_2^2
\end{multline}
following the ``lazy backtracking'' strategy as it was proposed for the iPiano
algorithm in~\cite{OCBP2013}. By testing for the validity of this inequality
also very small Lipschitz constants may be accepted, if the new value
$f(\tilde{\vec{z}}^{(\ell+1)})$ is even lower than what would be possible for a
function $f$ with an $L^{(\ell)}$-Lipschitz continuous gradient, for more
details see also Section~\ref{sec:aspects-of-iPiano}.
%
\subsection{Descent properties of the iPiano algorithm}
\label{sec:descent-properties}
We have seen in Section~\ref{sec:gradientApprox} that it is not guaranteed 
that the approximated gradient $\vec{q}$ delivers a descent direction for the
function $f(\vec{z})$. Testing if $-\vec{q}(\vec{z})$ is a descent direction
could be done by computing the actual gradient $\nabla f (\vec{z})$, which is
not desirable for practical applications.\par
Another test may be to watch for increasing energies
$f(\tilde{\vec{z}}^{(\ell)})+g(\tilde{\vec{z}}^{(\ell)})$ during computations performed by the iPiano
algorithm. However, iPiano does not enforce decreasing function values, but a
descent property is given for a majorising sequence of values
\begin{equation}
H_{\delta^{(\ell)}}(\tilde{\vec{z}}^{(\ell)},\tilde{\vec{z}}^{(\ell-1)})
\coloneqq f(\tilde{\vec{z}}^{(\ell)})+g(\tilde{\vec{z}}^{(\ell)})+\delta^{(\ell)}\Delta^{(\ell)}
\end{equation}
as pointed out in~\cite{OCBP2013}, Proposition~4.7, where
\begin{align}
\Delta^{(\ell)}&\coloneqq\norm*{\tilde{\vec{z}}^{(\ell)}-\tilde{\vec{z}}^{(\ell-1)}}_2^2\\
\delta^{(\ell)}&\coloneqq \frac{1}{\alpha^{(\ell)}}-\frac{L^{(\ell)}}{2}-\frac{\beta^{(\ell)}}{\alpha^{(\ell)}}
\end{align}
For sequences $\lbrace \tilde{\vec{z}}^{(\ell)}\rbrace_{\ell=-1}^{\infty}$,
$\lbrace L^{(\ell)}\rbrace_{\ell=0}^{\infty}$,
$\lbrace \alpha^{(\ell)}\rbrace_{\ell=0}^{\infty}$ and
$\lbrace \beta^{(\ell)}\rbrace_{\ell=0}^{\infty}$ generated by iPiano, the
sequence
$\lbrace
H_{\delta^{(\ell)}}(\tilde{\vec{z}}^{(\ell)},\tilde{\vec{z}}^{(\ell-1)})\rbrace_{\ell=0}^{\infty}$
is monotonically decreasing, and for $\ell=0,1,\dots$,
\begin{equation}
      H_{\delta^{(\ell+1)}}(\tilde{\vec{z}}^{(\ell+1)},\tilde{\vec{z}}^{(\ell)})\leq
      H_{\delta^{(\ell)}}(\tilde{\vec{z}}^{(\ell)},\tilde{\vec{z}}^{(\ell-1)})-\gamma^{(\ell)}\Delta^{(\ell)}
      \label{eq:descent-property}
\end{equation}
holds, where
\begin{equation}
      \gamma^{(\ell)}\coloneqq \frac{1}{\alpha^{(\ell)}}-\frac{L^{(\ell)}}{2}-\frac{\beta^{(\ell)}}{2\alpha^{(\ell)}}
\end{equation}
While using the approximated gradient $\vec{q}(\vec{z})$ in our numerical
experiments (\cf~Section~\ref{sec:numerical-evaluation}), the
property~\eqref{eq:descent-property} always holds for $\ell> 0$. \par
In our numerical experiments we could sometimes observe increasing energies
$f(\tilde{\vec{z}}^{(\ell)})+g(\tilde{\vec{z}}^{(\ell)})$, but they were always accompanied by decreasing
distances $\Delta^{(\ell)}$, leading to a convergent state. In these cases the
energies in the convergent state are always lower than the initial energy
$f(\tilde{\vec{z}}^{(0)})+g(\tilde{\vec{z}}^{(0)})$.\par
When using the exact gradient $\nabla f$, we did not observe increasing energy
values in our experiments. However, since the computation of $\vec{q}$ is a lot
faster and we could achieve good results with the approximated gradient, we
regard $\vec{q}$ as a more efficient approximation of $\nabla f$.\par
For a constant step size $\beta^{(\ell)}=\beta$, the sequence $\lbrace
\delta^{(\ell)}\rbrace_{l=0}^{\infty}$ may not be monotonically decreasing, so
that the convergence theory provided in~\cite{OCBP2013} may not be applicable.
This can be fixed be employing a variable $\beta^{(\ell)}$ in
Algorithm~\ref{alg:iPianoPS}, following the proof of Lemma~4.6
in~\cite{OCBP2013}.\par
Thus for every $\ell$ we compute the auxiliary variable $\nu \coloneqq
(\delta_{\ell-1}+\frac{L^{(\ell)}}{2})/(c+\frac{L^{(\ell)}}{2})$ and set
\begin{equation}
      \beta^{(\ell)}=\frac{\nu-1}{\nu-\frac{1}{2}+c}
\end{equation}
As initialisation we set $\delta^{(-1)}$=1.
%
\section{Numerical evaluation}
\label{sec:numerical-evaluation}
%
In this section we discuss some numerical experiments as well as important 
observations.\par
For all our experiments, the stopping criterion was set to a test on the
relative change in the objective function ($<10^{-8}$), evaluated on
$\tilde{\vec{z}}^{(\ell)}$ in the inner iPiano loop and on $\vec{z}^{(k)}$ in
the outer loop. Note that in the outer loop different albedos are used, \ie~the
energies $f(\vec{z}^{(k)},\rho^{(k)})+g(\vec{z}^{(k)})$ and
$f(\vec{z}^{(k+1)},\rho^{(k+1)})+g(\vec{z}^{(k+1)})$ are being evaluated. Also
the maximum number of iterations was set to 100 in the inner iPiano loop and 500
in the outer loop, if not specified otherwise.\par
%
\subsection{Computational aspects of iPiano}
\label{sec:aspects-of-iPiano}
\begin{figure}
      \centering
      \begin{tabular}{cc}
        \includegraphics{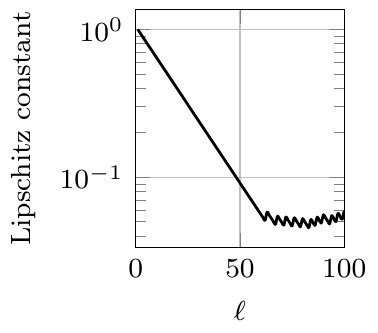} & \includegraphics{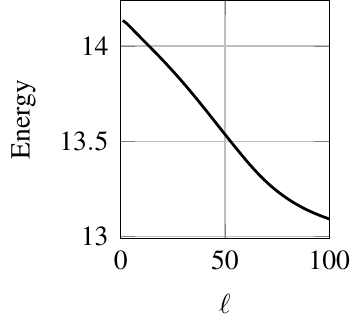} \\
        (a) & (b)
      \end{tabular}
      \caption{The \emph{Cat} experiment, $k=1$:
	(a) local Lipschitz constants obtained by the lazy backtracking strategy;
	(b) objective function $f+g$ as a function of the iPiano iterations count $\ell$.
	\label{fig:Lipschitz_cat1}}
\vspace{4pt}
      \begin{tabular}{cc}
        \includegraphics{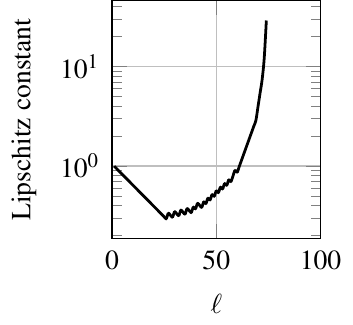} & \includegraphics{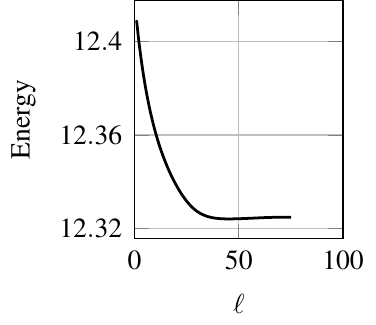} \\
        (a) & (b)
      \end{tabular}
      \caption{The \emph{Cat} experiment, $k=2$:
        (a)~local Lipschitz constants obtained by the lazy-backtracking strategy. At $\ell=75$
        the increase generates a small $\alpha^{(\ell)}$, therefore the iPiano break criteria is fulfilled;
        (b)~objective function $f+g$ as a function of the iPiano iterations count~$\ell$. Starting at $\ell=46$ 
        the objective function slightly increases. This is not contrary to the convergence theory, since the 
        descent property is fulfilled for a majorising sequence.
        \label{fig:Lipschitz_cat2}}
\vspace{4pt}
      \begin{tabular}{cc}
        \includegraphics{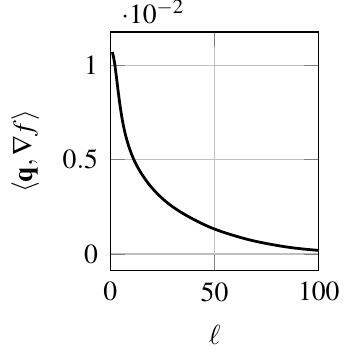} & \includegraphics{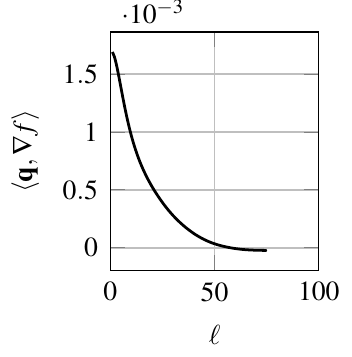} \\
        (a) & (b)
      \end{tabular}
      \caption{The \emph{Cat} experiment: $\langle \textbf{q}, \nabla f \rangle$ for
	(a) $k=1$ and (b) $k=2$. For non-negative values the vector $-\vec{q}(\vec{z}^{(\ell)})$
	is a descent direction.
	\label{fig:approx}}
\end{figure}

%
Let us first recall that for a function $f:\Real^{n}\rightarrow \Real$ with a
Lipschitz continuous gradient, such that
\begin{equation}
      \norm*{\nabla f(\vec{x})-\nabla f(\vec{y})}_2
      \leq L^{\nabla f}\norm*{\vec{x}-\vec{y}}_2
\end{equation}
for all $\vec{x}$, $\vec{y}\in\Real^{n}$ we have
\begin{equation}
      \abs*{f(\vec{x})-f(\vec{y})-\left\langle\nabla f(\vec{y}),\vec{x}-\vec{y}\right\rangle}
      \leq \frac{L^{\nabla f}}{2}\norm*{\vec{x}-\vec{y}}_2^2
\end{equation}
for all $\vec{x}$, $\vec{y}\in\Real^{n}$, see \eg~\cite{OR1970} Theorem in 3.2.12. This
leads to the property
\begin{equation}
      \label{eq:descent-lemma}
      f(\vec{x})\leq f(\vec{y})+\left\langle\nabla f(\vec{y}),\vec{x}-\vec{y}\right\rangle
      + \frac{L^{\nabla f}}{2}\norm*{\vec{x}-\vec{y}}_2^2,
\end{equation}
for all $\vec{x}$, $\vec{y}\in\Real^{n}$, which is also the subject of the
descent lemma, \cf~\cite{OCBP2013}~Lemma~4.1. In the iPiano algorithm only the necessary
condition~\eqref{eq:descent-lemma} is tested with
$\vec{x}=\tilde{\vec{z}}^{(\ell+1)}$ and $\vec{y}=\tilde{\vec{z}}^{(\ell)}$ and
used to derive a local Lipschitz constant. By this, one can allow step sizes
leading to a steeper (better) descent in $f$. In our experiments we often
encountered rather low local Lipschitz constants, some examples are depicted in
Fig.~\ref{fig:Lipschitz_cat1}. Sometimes we encountered increasing local Lipschitz 
constants towards the end of an iPiano instance. These would then lead to decreasing 
step sizes $\alpha^{(\ell)}$, such that finally the break criteria for the iPiano
algorithm would be fulfilled. An example is depicted in Fig.~\ref{fig:Lipschitz_cat2} (a).\par
While in most iterates in our experiments the energy $f(\vec{z})+g(\vec{z})$ was
decreasing, sometimes it was slightly increasing towards the end oft the
sequence of iPiano iterations, see Fig.~\ref{fig:Lipschitz_cat2} (b). We
conjecture that this is related to approximated gradients $\vec{q}$, which do
not deliver a descent direction with $\langle\vec{q},\nabla f\rangle\geq 0$, see
also Fig.~\ref{fig:approx}.\par
We did not observe any spikes in the sequence of local Lipschitz or increasing
energies when the exact gradient~\eqref{eq:grad_f} was used. Therefore the use
of the exact gradient would lead to a somehow smoother and faster convergence in
terms of number of iterations. However, $\vec{q}(\vec{z})$ can be computed much
faster than $\nabla f(\vec{z})$ and we did not observe the exact gradient
leading to local minima of~\eqref{eq:19} with significantly smaller energies, so
that we still regard the use of the approximated gradient as the more feasible
alternative. In detail, the average computation time (over 100 evaluations) of
the exact gradient is roughly 55 seconds, whereas the simplified gradient can be
evaluated in 0.13 seconds, which results in a speedup factor of more than 400.
%
\subsection{Numerical results}
\label{sec:numerical-results}
\begin{figure*}
      \centering
      \begin{tabular}{ccccc}
        \includegraphics[width=0.175\textwidth]{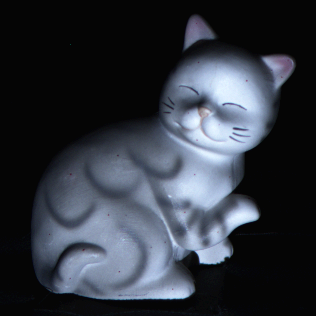}   &
        \includegraphics[width=0.175\textwidth]{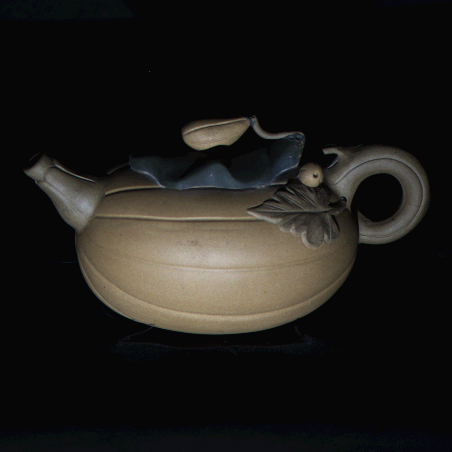}  &
        \includegraphics[width=0.175\textwidth]{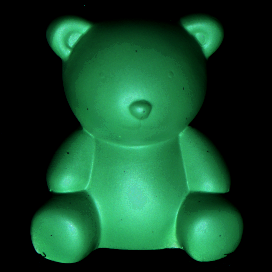}  &
        \includegraphics[width=0.175\textwidth]{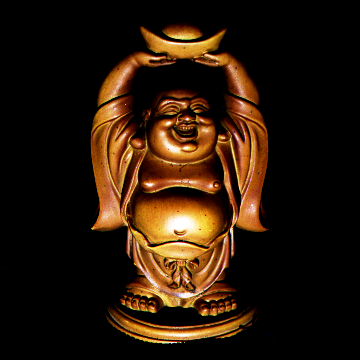}&
        \includegraphics[width=0.175\textwidth]{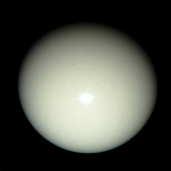}
        \\
        \includegraphics[width=0.175\textwidth]{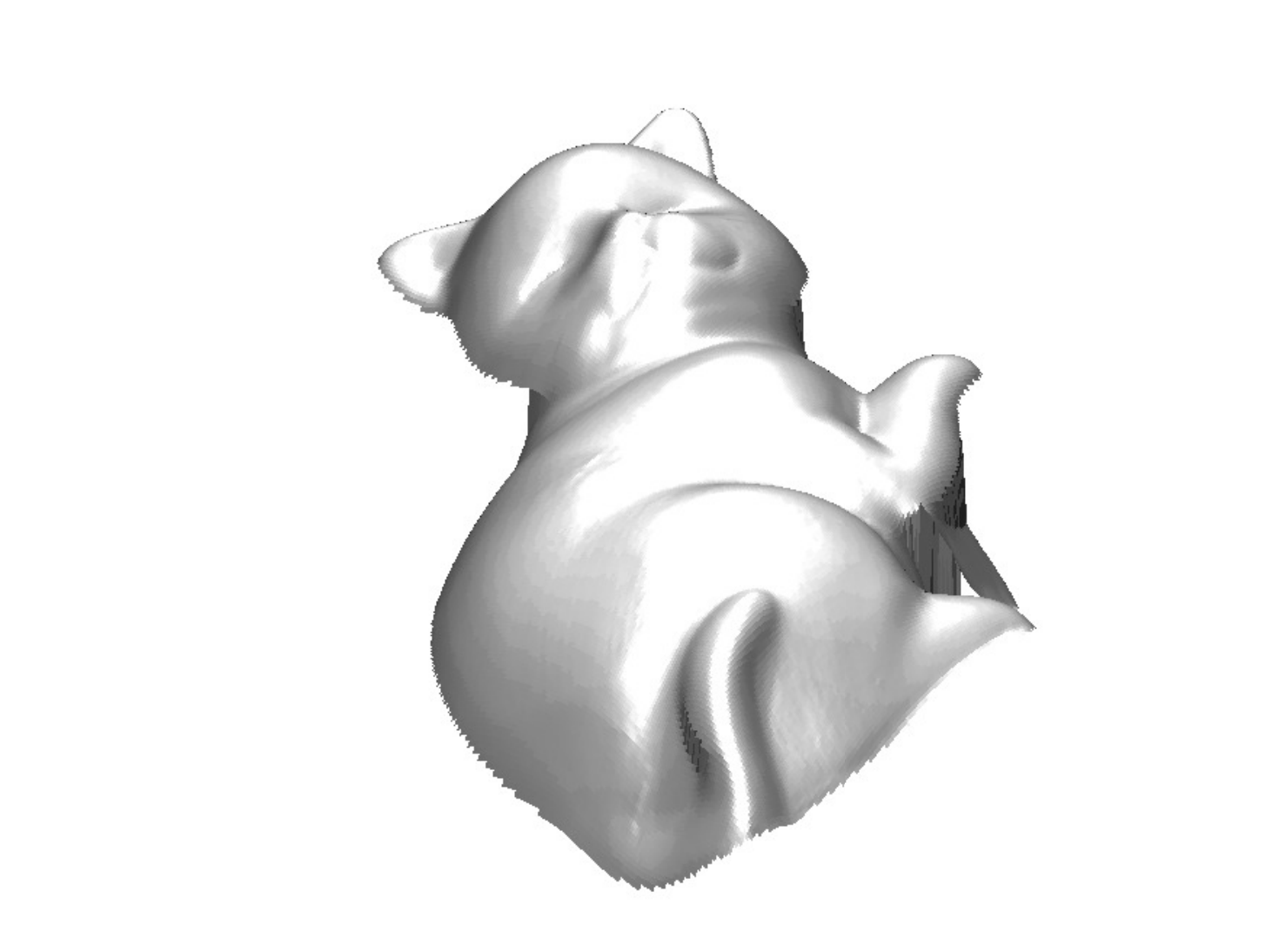} &
        \includegraphics[width=0.175\textwidth]{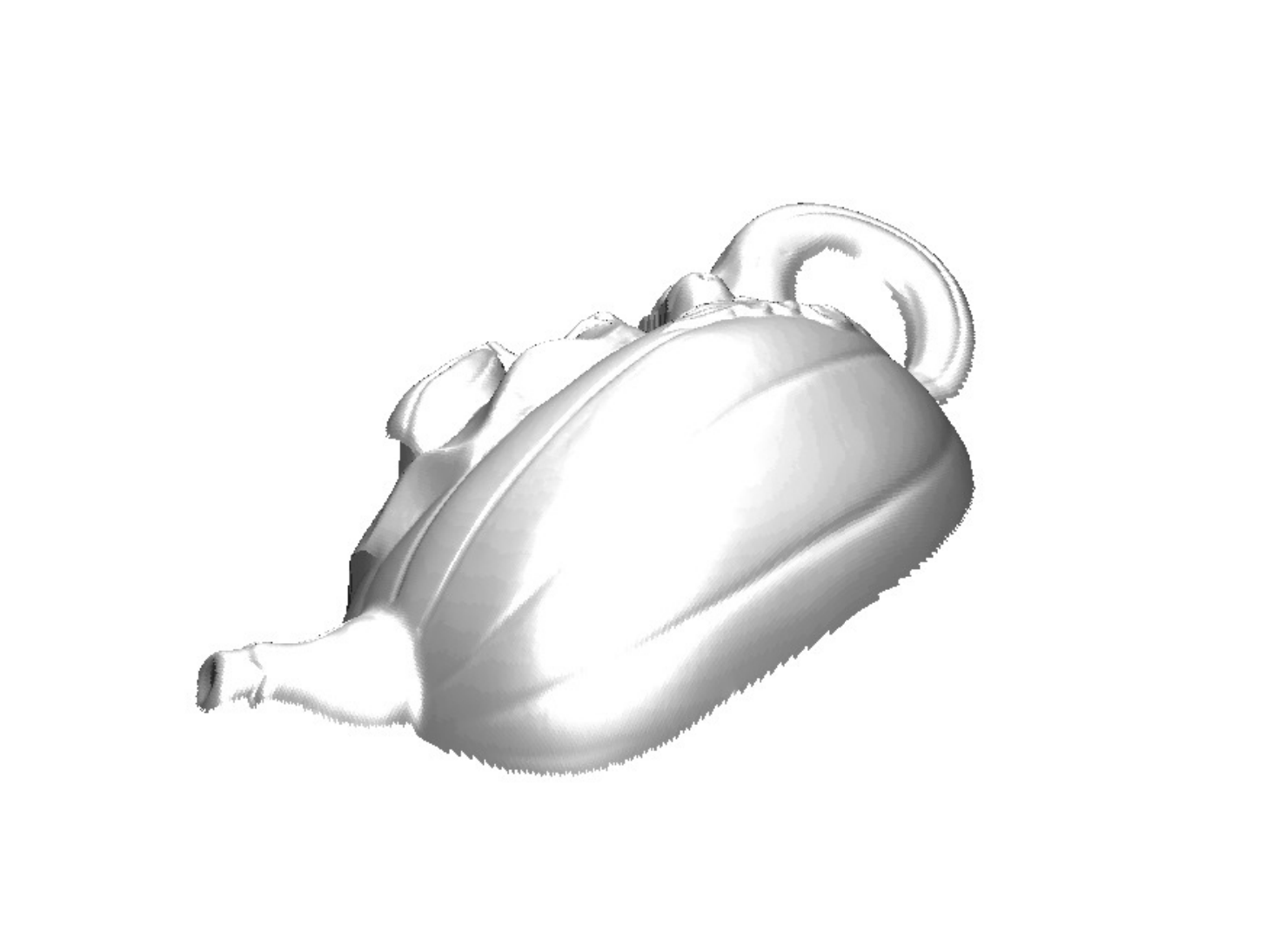} &
        \includegraphics[width=0.175\textwidth]{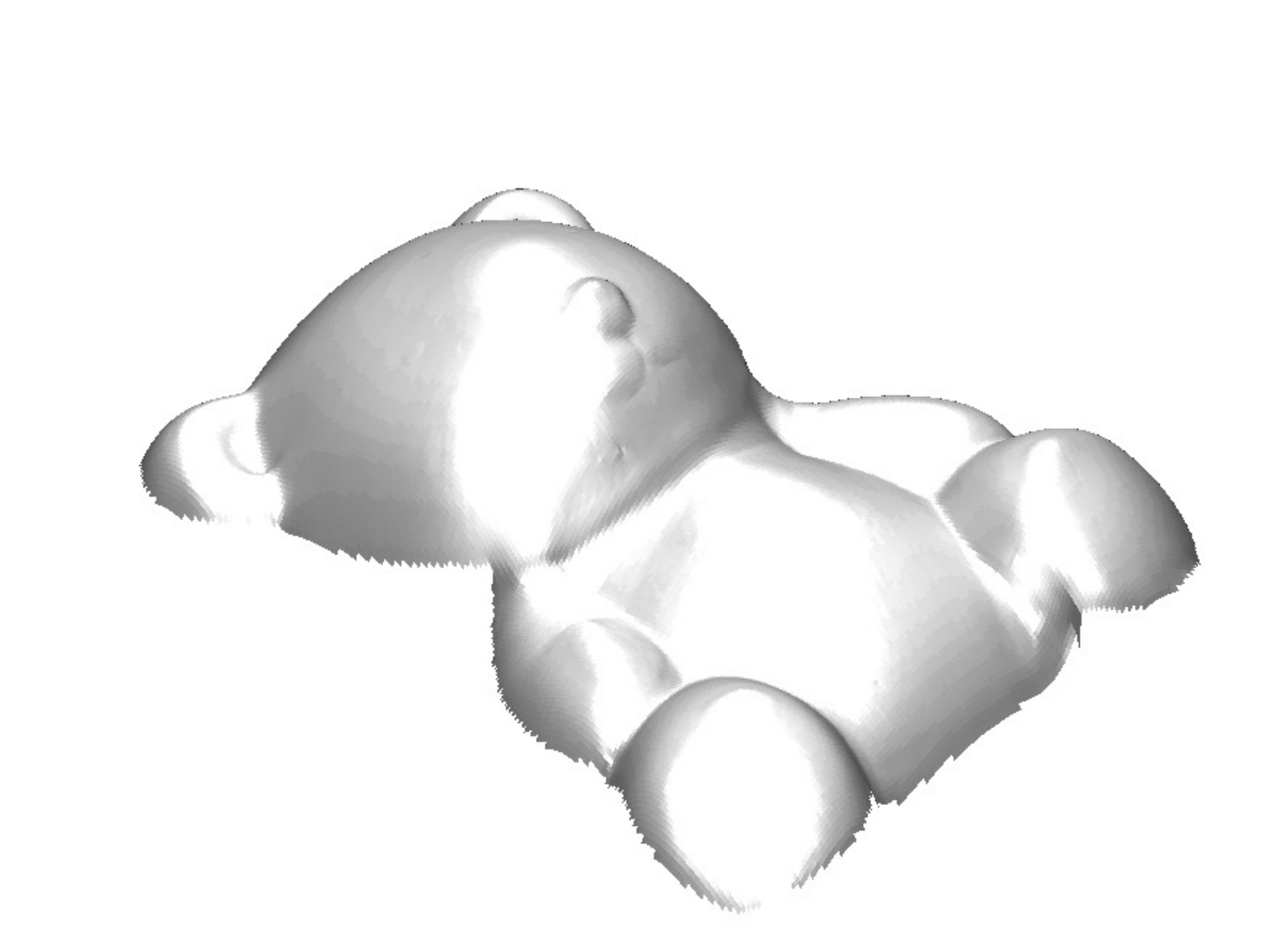} &
        \includegraphics[width=0.175\textwidth]{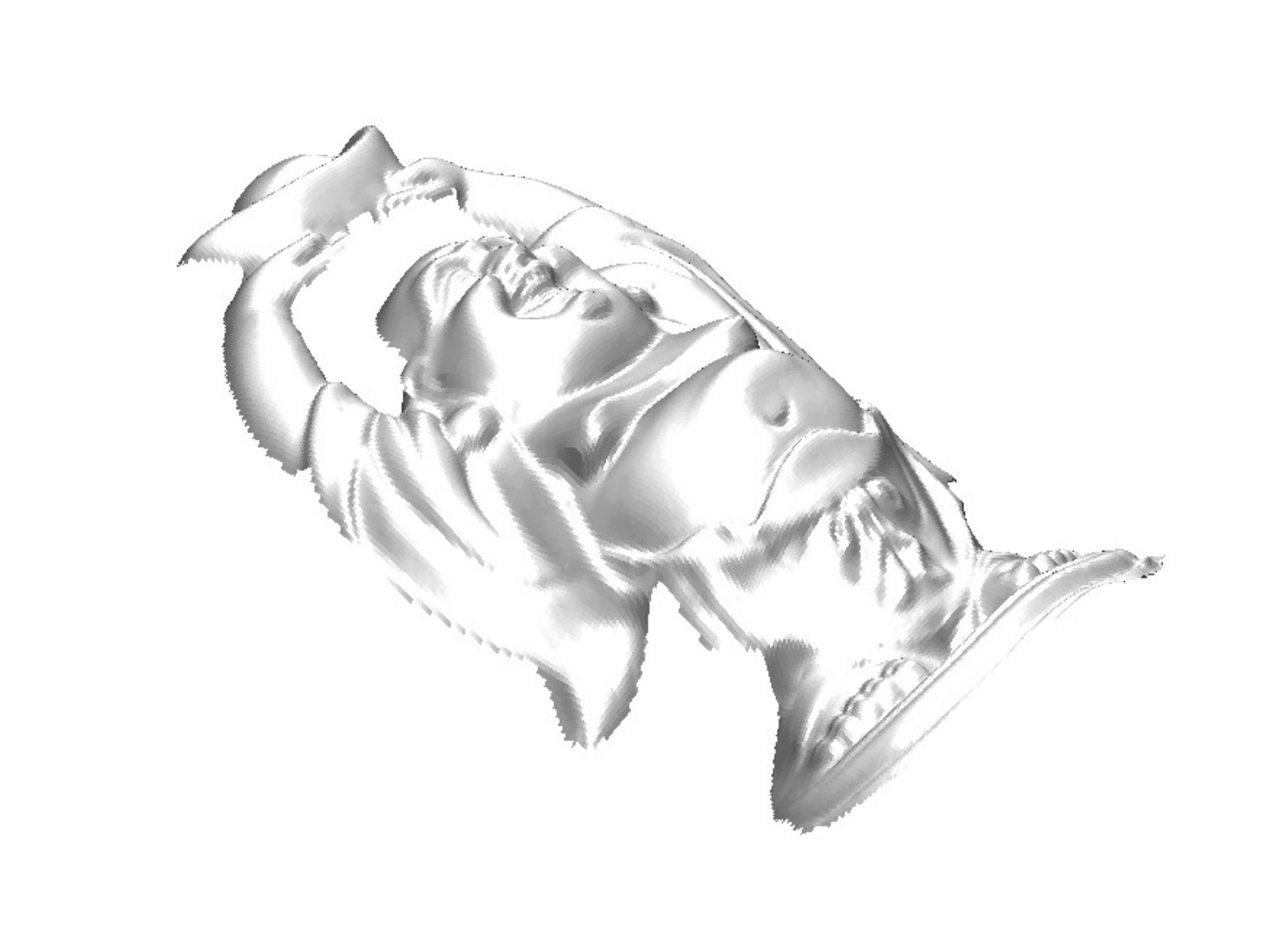} &
        \includegraphics[width=0.175\textwidth]{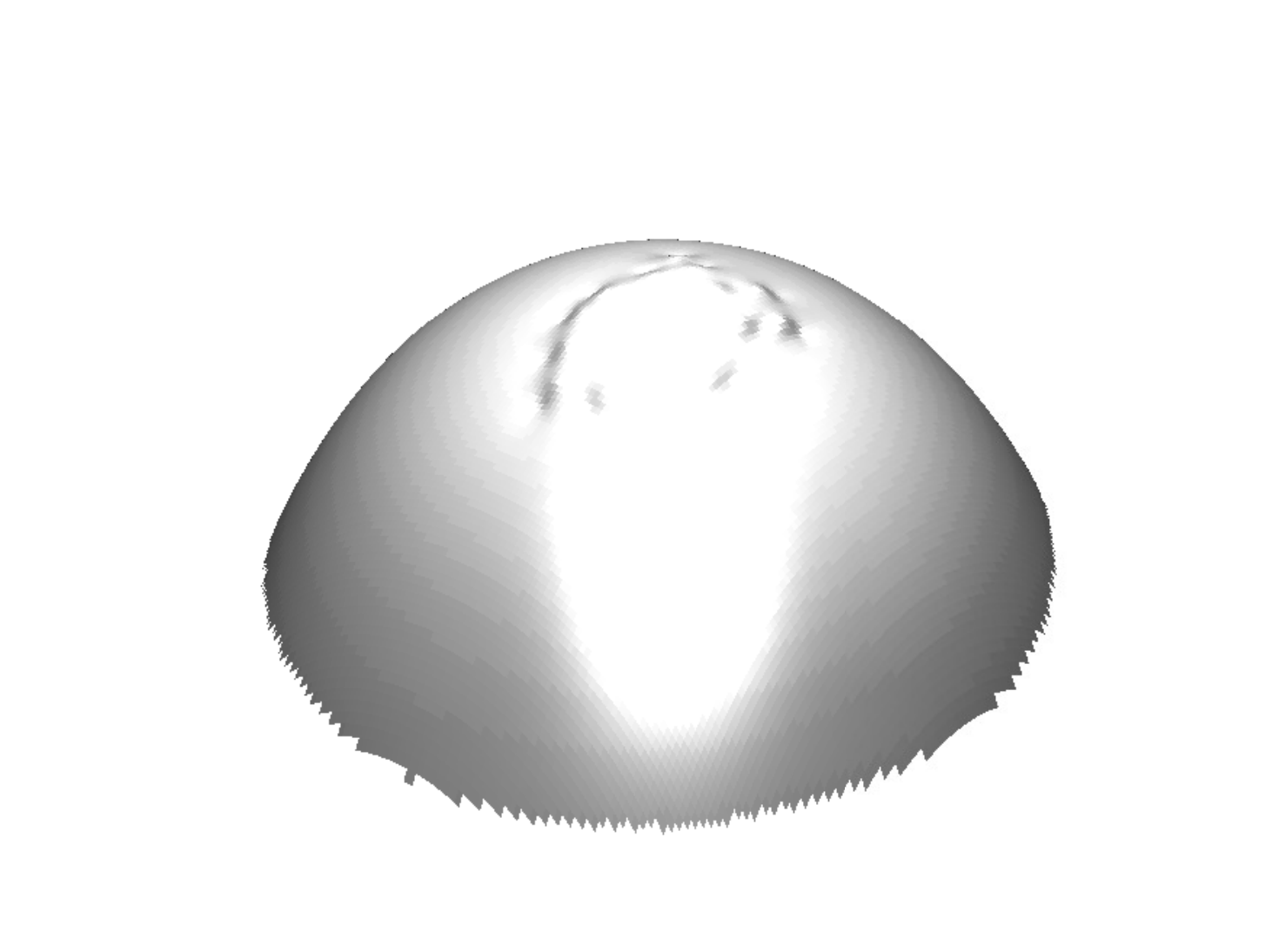}
        \\
        \emph{Cat} & \emph{Pot} & \emph{Bear} & \emph{Buddha} & \emph{Ball}
      \end{tabular}
      \caption{Test data (brightened and cropped to enhance visualisation) and
        3D-reconstructions obtained after $500$ iterations $k$ of
        Algorithm~\ref{alg:iPianoPS}.}
      \label{fig:TestData}
\end{figure*}

\begin{figure*}
	\centering
	\begin{tabular}{cccc}
		\includegraphics{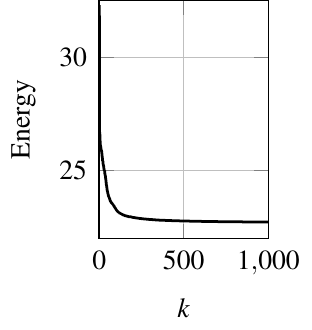}          &
		\includegraphics{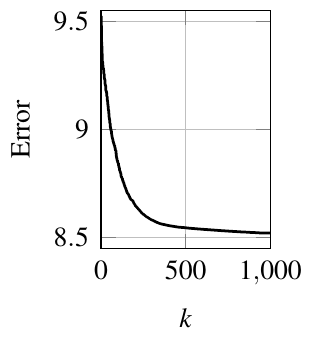}          &
		\includegraphics{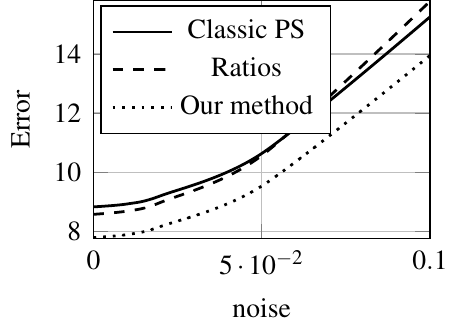}  &
		\includegraphics{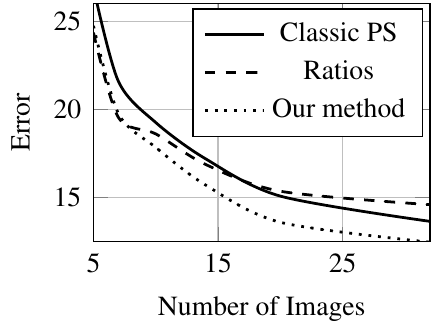} \\
		\hspace*{9mm}(a) & \hspace*{6mm}(b) & \hspace*{6mm}(c) & \hspace*{7mm}(d)
	\end{tabular}
	\vspace*{0.01ex}
	\caption{ The \emph{Cat} experiment:
		(a) objective function $f+g$ as a function of the iterations count $k$;
		(b) \gls{MAE} between the reconstructed surface and the ground
		truth; (c) reprojection error for competing methods for
		increasing noise levels (we indicate the standard deviation of the
		additive, zero-mean Gaussian noise, as a percentage of the maximum
		intensity); (d) ditto for increasing numbers of input images, with
		$0.1$ noise level.
		\label{fig:CV_graphs}}
\end{figure*}

%
Figure~\ref{fig:TestData} presents the test data that we use in this paper. It
consists of five real-world scenes captured under 20 different known illuminants
$\vec{s}^{i}$, ($i=1$, \ldots, $20$), provided in~\cite{Shi2016}. In our
experiments, we used $m=20$ out of the original $96$ RGB images, which we
converted to grey levels. Two of the sets present diffuse reflectance
(\emph{Cat} and \emph{Pot}), while two other exhibit broad specularities
(\emph{Bear} and \emph{Buddha}) and one presents sparse specular spikes
(\emph{Ball}). Since the ground truth normals are also provided
in~\cite{Shi2016}, the estimated normals can be computed from the final depth
map according to~\eqref{eq:def_normal}, and compared to the exact ground truth.
For evaluation, we indicate the \gls{MAE} (in degrees) over the reconstruction
domain $\Omega$.
Let us consider the \emph{Cat} data set in some detail, as it consists of a
diffuse scene that fits rather well our assumptions. We let our algorithm run
for $1000$ outer iterations $k$ (approx.\ $1$ hour on a recent i7 processor, using
non-optimised Matlab code), and study the evolution of two criteria: the
reprojection error, whose minimum is sought by our algorithm; and the \gls{MAE},
which indicates the overall accuracy of the 3D reconstruction, \cf~the two left
images within Figure \ref{fig:CV_graphs}. The displayed convergence graphs
indicate that each iteration from Algorithm~\ref{alg:iPianoPS} not only
decreases the value of the objective function $f+g$ (which is approximately
equal to the reprojection error $\mathcal{E}_{\mathcal{R}} = f$), but also the
\gls{MAE}. This confirms our conjecture that finding the best possible
explanation of the images yields more accurate 3D reconstructions.
In the other two graphs in Figure \ref{fig:CV_graphs} we study the results of
our method compared to other \gls{PS} strategies based on least-squares: the
classical \gls{PS} framework~\cite{Woodham1980} consisting in estimating in a
least squares sense the normals and the albedo, and integrating them afterwards,
and the recent differential ratios procedure from~\cite{Mecca2016}, forcing 
Lambertian reflectance and least-squares estimation, for fair comparison. The
latter allows direct recovery of the depth, but on the other hand it changes the
objective function. Both other approaches rely on linear least squares: they are
thus by far faster than the proposed approach (here, a few seconds, versus a few
minutes with ours). Yet, in terms of accuracy, these methods are outperformed by
our approach, no matter the noise level or the number of images (which were
preprocessed via low-rank factorisation~\cite{Wu2010} in these two
experiments).\par
\begin{figure*}
      \centering
      \begin{tabular}{ccccc}
        \includegraphics[width = 0.175\textwidth]{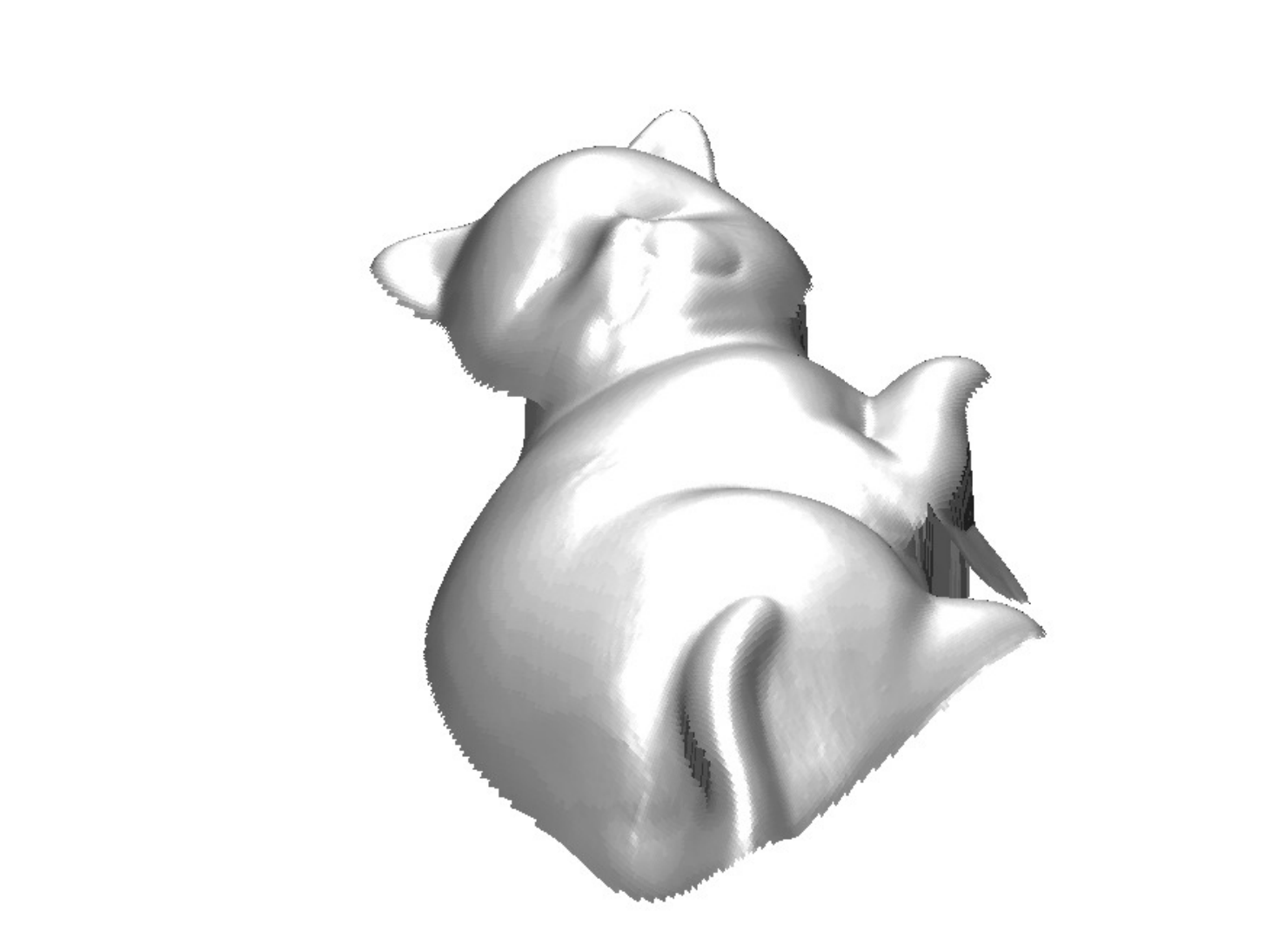} &
        \includegraphics[width = 0.175\textwidth]{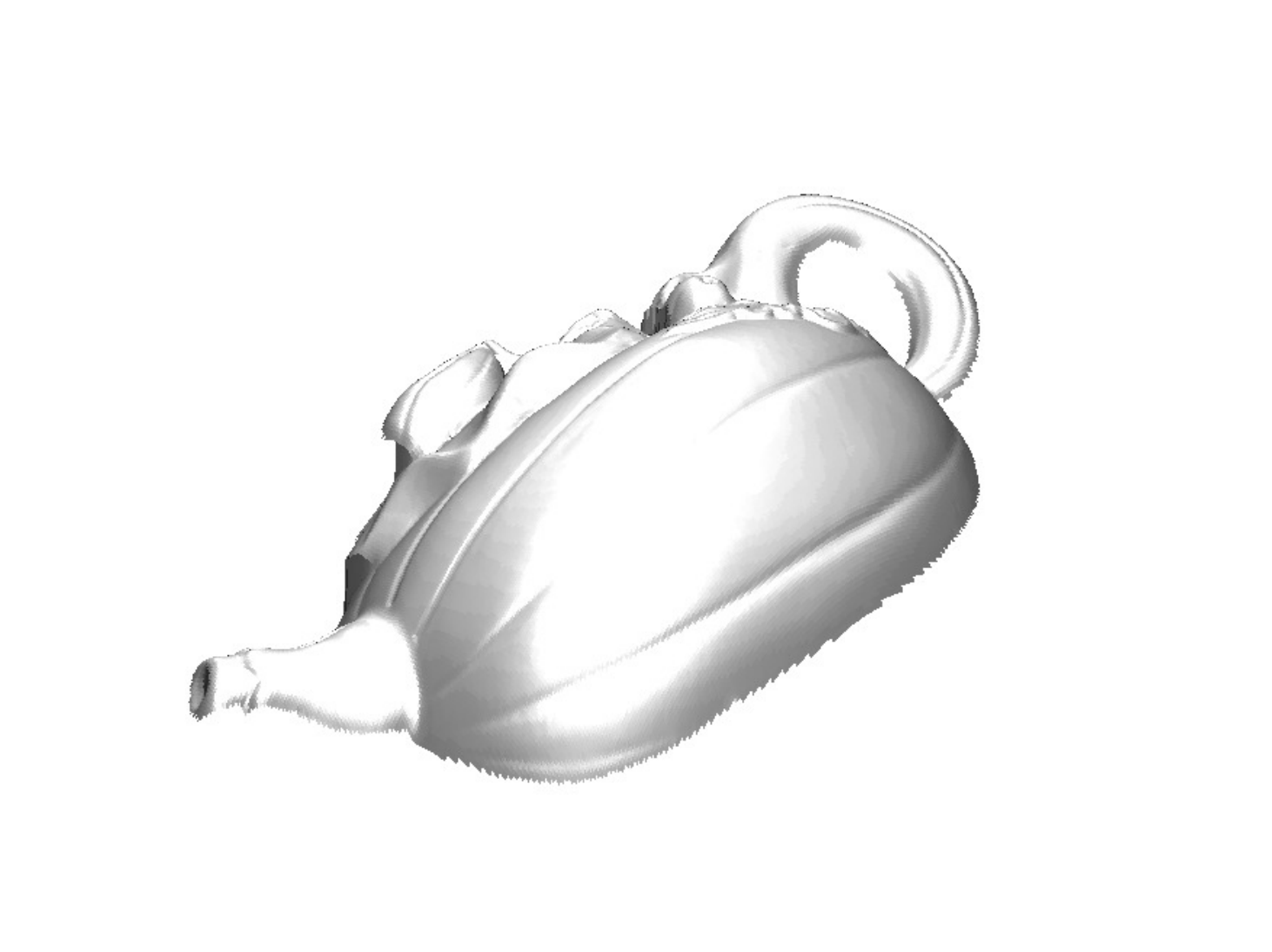} &
		\includegraphics[width = 0.175\textwidth]{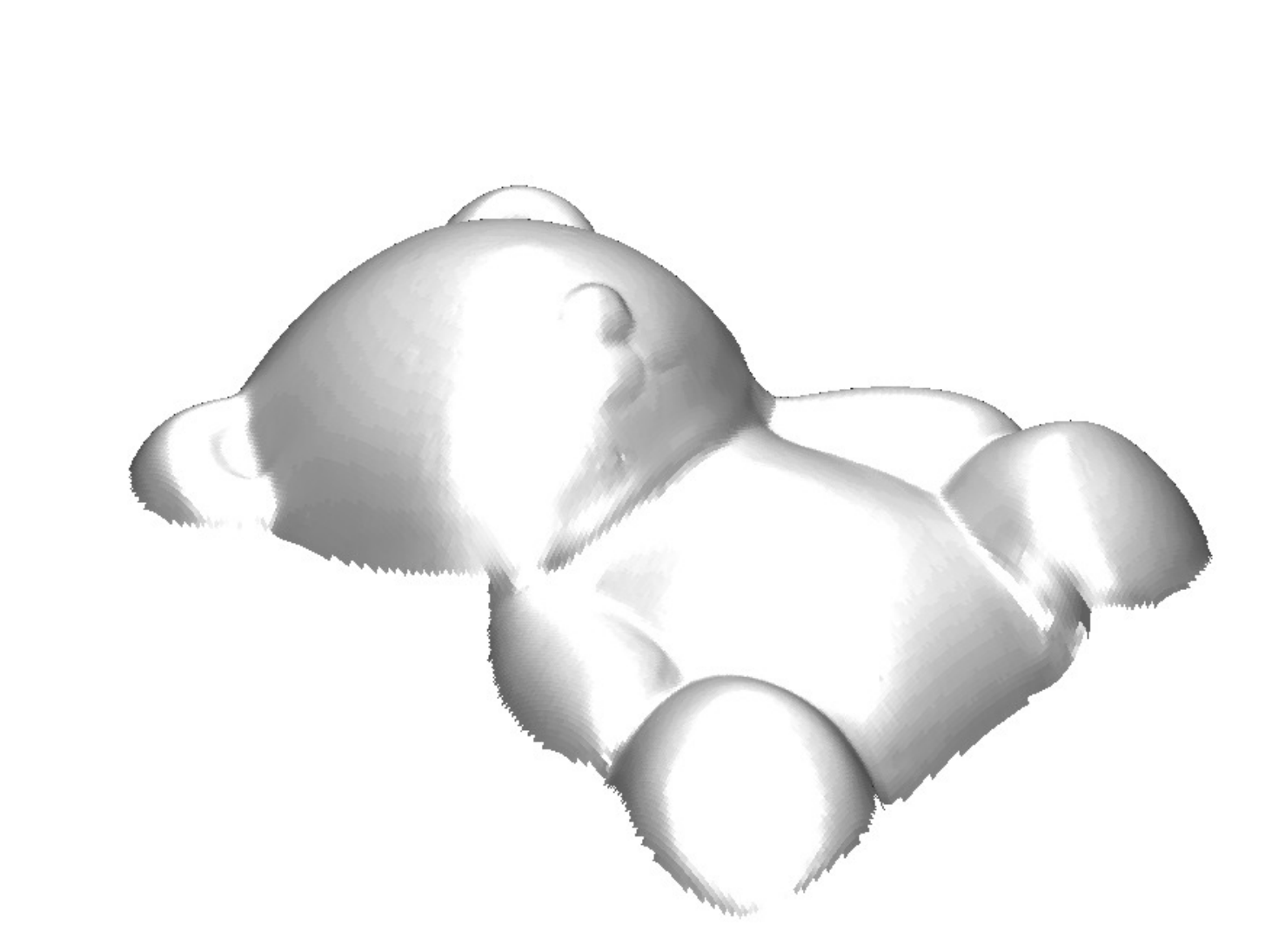} &
		\includegraphics[width = 0.175\textwidth]{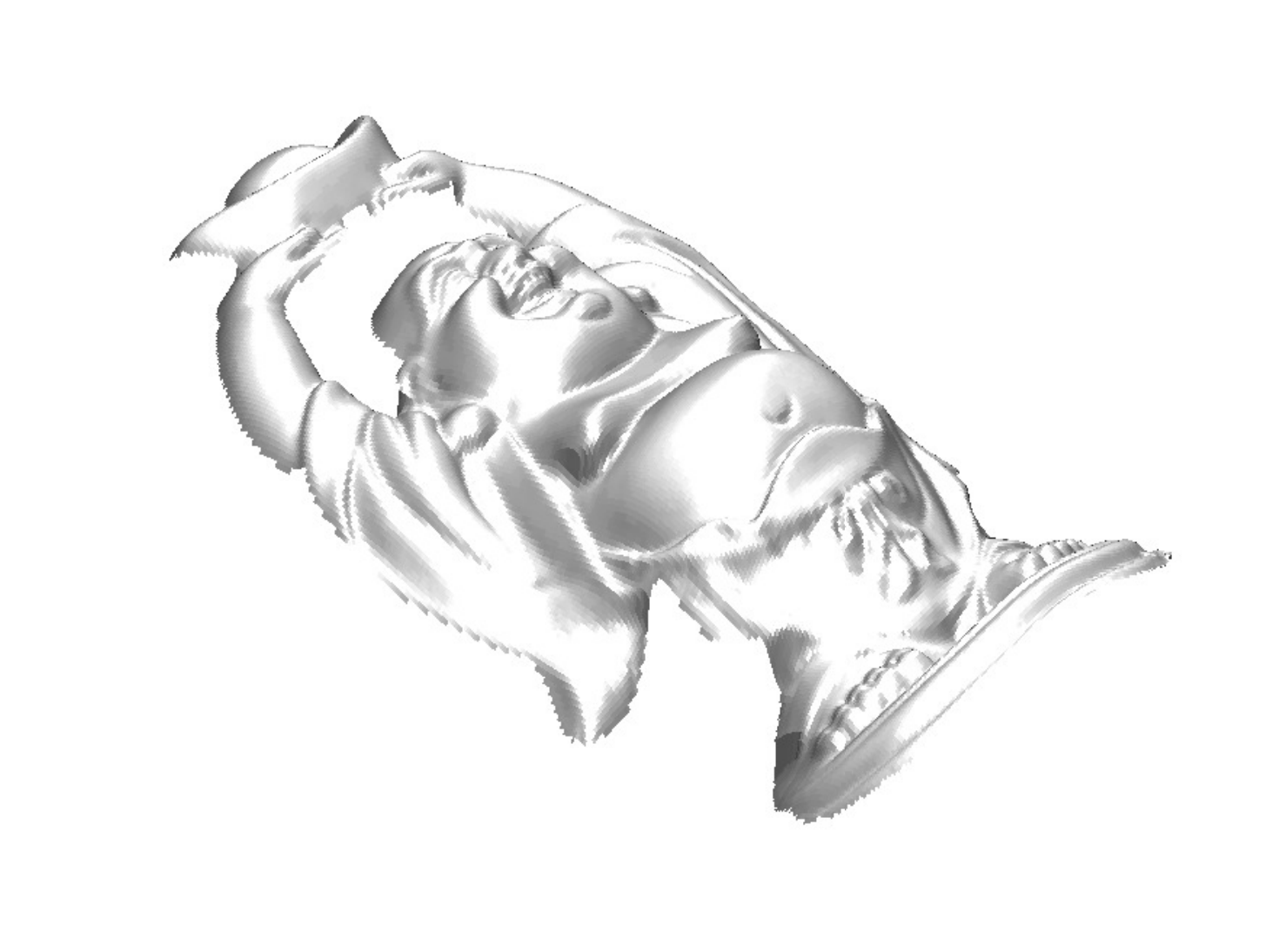} &
		\includegraphics[width = 0.175\textwidth]{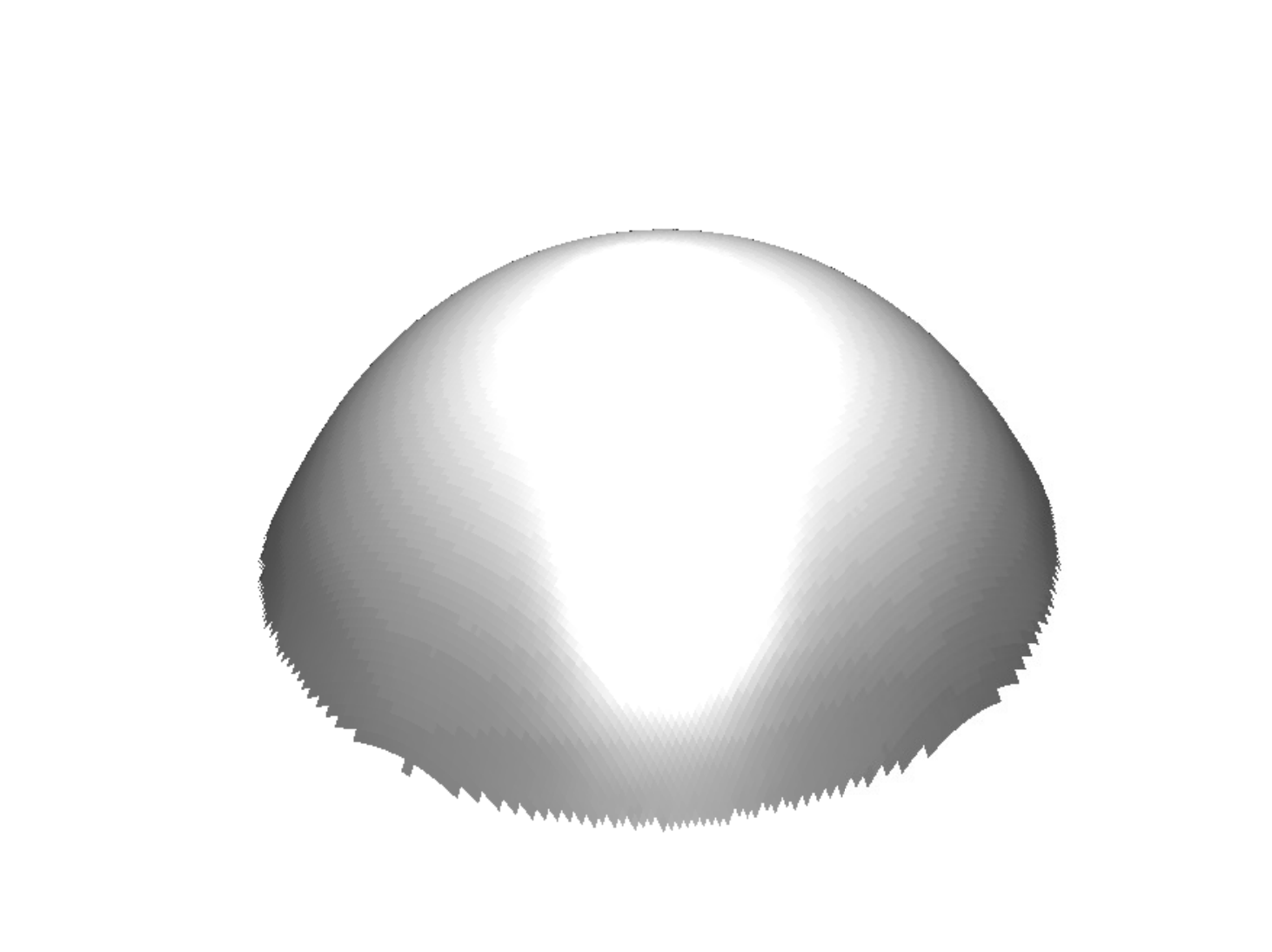}
      \end{tabular}
      \caption{3D-reconstruction results using the \emph{full pipeline},
        consisting of a preprocessing~\cite{Wu2010}, followed by classic
        \gls{PS}~\cite{Woodham1980}, and finally the proposed method.}
	\label{fig:3D_robust}
\end{figure*}

By making the input images Lambertian via low-rank preprocessing~\cite{Wu2010},
we can make a reasonable comparison for the whole test dataset.
Table~\ref{tab:comparison2} shows that our postprocessing method can still
improve the accuracy. The 3D re\-con\-struction results obtained with the full
pipeline are shown in Figure~\ref{fig:3D_robust}. In comparison with
Figure~\ref{fig:TestData}, artefacts due to specularities are clearly reduced.
\begin{table}
	\centering
	\caption{Reconstruction errors (\gls{MAE}, in degrees) obtained for
          preprocessed input images using the approach from~\cite{Wu2010}. For
          fair comparison, the \gls{MAE} for classic \gls{PS} is calculated on
          the final surface, \ie~using the normals calculated by finite
          differences from the final depth map, rather than the (non-integrable)
          normals estimated in the first step. Regarding the ratio procedure, we
          applied the code from~\cite{Mecca2016} directly on the grey level
          data.}
	\begin{tabular}{l*{5}{c}}
		\addlinespace \toprule \addlinespace
		& Cat &  Pot & Bear & Buddha & Ball \\
		\cmidrule(r){1-1} \cmidrule(l){2-6}
		Classic \gls{PS}~\cite{Woodham1980} & 8.83 & 8.92 & 7.01 & 14.34 & 3.05 \\
		Differential ratios~\cite{Mecca2016} & 8.57 &  9.00 & 7.01 & 14.31 & 3.13 \\
		Our method (500 iter.) & \textbf{7.79} &  \textbf{8.58} & \textbf{6.90} & \textbf{13.89} & \textbf{2.97} \\
		\addlinespace \bottomrule
	\end{tabular}
	\label{tab:comparison2}
\end{table}
%
\section{Conclusions}
\label{sec:conclusions}
%
We have shown the benefits of recent, high performing numerical methods in the
context of photometric stereo. Let us emphasise that only by considering such
recent developments in numerical optimisation methods complex models as arising
in \gls{PS} can be handled with success. Our results show that a significant
quality gain can be achieved in this way while at the same time the mathematical
proceeding can be validated rigorously.\par
Our experimental investigation has shown what can be expected from the basic
iPiano method as well as by computational simplifications as proposed by us in
terms of an approximated gradient. In particular we have shown that it may not
be easy to interprete relevant properties of computed iterates.\par
A more detailed view on the computational results reveals that remaining
inaccuracies seem to be mostly due to shadows and highlights, edges and depth
discontinuities. Thus, an interesting perspective of our work would be to use
more robust estimators, which would ensure both robustness to
outliers~\cite{Ikehata2014a,Queau2017} and improved preservation of
edges~\cite{Durou2009}.
\bibliographystyle{spmpsci}
\bibliography{finalversion-6}
%
\end{document}